\documentclass[journal]{IEEEtran}

\usepackage{amsfonts}
\usepackage{amssymb}
\usepackage{booktabs}
\usepackage[cmex10]{amsmath}
\usepackage{algorithm}
\usepackage[noend]{algpseudocode}
\usepackage{multirow}
\usepackage[switch]{lineno}
\usepackage{color}
\usepackage{graphicx}
\graphicspath{{./imgs/}{./figures}{./appendix_img/}}

\makeatletter
\let\NAT@parse\undefined
\makeatother
\usepackage{hyperref}  

\newcommand{\RmNum}[1]{\uppercase\expandafter{\romannumeral#1}} 
\newcommand{\revise}[1]{#1} 

\begin{document}

\title{FACEMUG: A Multimodal Generative and Fusion Framework for Local Facial Editing}

\author{Wanglong Lu, Jikai Wang, Xiaogang Jin, Xianta Jiang, and Hanli Zhao$^{*}$
\thanks{W. Lu, J. Wang, and H. Zhao are with the College of Computer Science and Artificial Intelligence, Wenzhou University, Wenzhou 325035, China.}
\thanks{W. Lu and X. Jiang are with the Department of Computer Science, Memorial University of Newfoundland, St. John's, NL A1B 3X5, Canada.}
\thanks{X. Jin is with the State Key Laboratory of CAD\&CG, Zhejiang University, Hangzhou 310058, China.}
\thanks{$^{*}$ Corresponding author. E-mail: hanlizhao@wzu.edu.cn}
}



\maketitle

\begin{abstract}
Existing facial editing methods have achieved remarkable results, yet they often fall short in supporting multimodal conditional local facial editing. One of the significant evidences is that their output image quality degrades dramatically after several iterations of incremental editing, as they do not support local editing. In this paper, we present a novel multimodal generative and fusion framework for globally-consistent local facial editing (FACEMUG) that can handle a wide range of input modalities and enable fine-grained and semantic manipulation while remaining unedited parts unchanged. Different modalities, including sketches, semantic maps, color maps, exemplar images, text, and attribute labels, are adept at conveying diverse conditioning details, and their combined synergy can provide more explicit guidance for the editing process. We thus integrate all modalities into a unified generative latent space to enable multimodal local facial edits. Specifically, a novel multimodal feature fusion mechanism is proposed by utilizing multimodal aggregation and style fusion blocks to fuse facial priors and multimodalities in both latent and feature spaces. We further introduce a novel self-supervised latent warping algorithm to rectify misaligned facial features, efficiently transferring the pose of the edited image to the given latent codes. We evaluate our FACEMUG through extensive experiments and comparisons to state-of-the-art (SOTA) methods. The results demonstrate the superiority of FACEMUG in terms of editing quality, flexibility, and semantic control, making it a promising solution for a wide range of local facial editing tasks.
\end{abstract}


\IEEEPARstart{T}{he} rapid development of digital imaging and mobile computing has fueled the demand for personalized content in social media and various applications~\cite{jiang_tvcg_2023,Qu_tvcg_2023,Han_tvcg_2020}, making facial image editing an essential research area in computer graphics and computer vision. Faces are universally acknowledged as the most representative and expressive aspect of human beings, which makes facial editing a challenging task~\cite{Wang_tvcg_2023}.

Many image editing tools provide convenient guidance information to allow users to edit facial features interactively~\cite{Su_tvcg_2023}. In order to provide convenient user interfaces, many face image editing tools make use of various input modalities to guide the editing of facial features. In recent years, multimodal facial image editing has attracted considerable interest. Since multimodal models excel at conveying various types of conditioning information, their combined synergy can offer clearer descriptions to aid in facial editing. For example, semantics can define a facial image's coarse layout; sketches can detail its structure and texture; and text or attribute labels can adjust facial attributes, to name just a few. The support of local editing capability is also important in a variety of image editing applications. Local editing enables users to edit local image regions in an incremental manner while keeping the contents of unedited background regions unchanged.

\begin{figure*}
	\includegraphics[width=\textwidth]{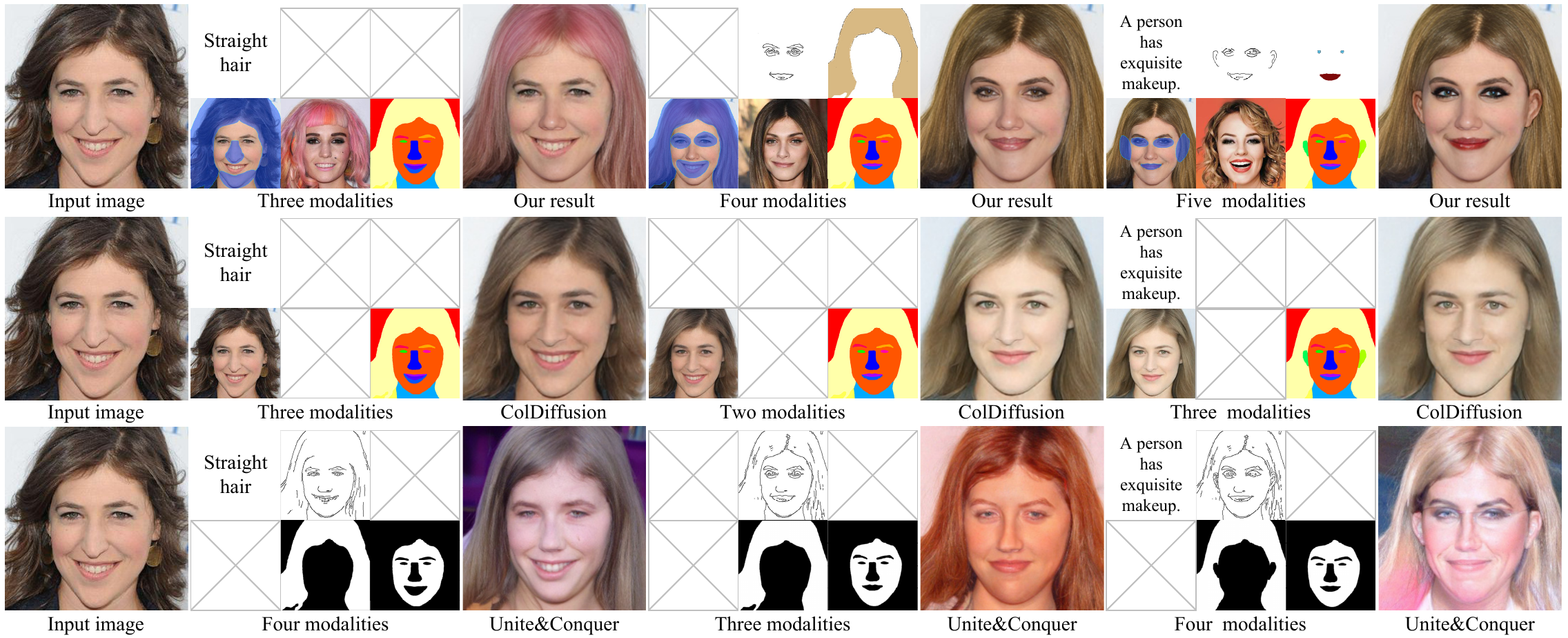}
     \caption{
     Examples demonstrating the superior performance of FACEMUG in high-quality globally consistent local facial editing, using subsets of the five modalities including semantic label, sketch, text, color, and exemplar image. Our method (top row) exhibits better visual quality and fidelity in incremental editing (the later editing taking the previous output image as input), compared to SOTA multimodal face editing methods: ColDiffusion~\cite{huang2023collaborative} (middle row) and Unite\&Conquer~\cite{nair2023unite} (bottom row).
     }
	\label{fig:fig_teaser_multi}
\end{figure*}

There are some limitations in existing multimodal local facial editing methods. The first limitation is the difficulty in maintaining visual contents in unedited background regions. Existing multimodal facial editing techniques~\cite{TediGAN_2021_CVPR,huang2022poegan,huang2023collaborative,nair2023unite} can only edit the facial image as a whole and are prone to introduce unwanted changes to unedited background regions. When users are not satisfied with some local effects, these techniques will fail to edit the local regions in an incremental manner.
As shown in Fig.~\ref{fig:fig_teaser_multi} (bottom two rows), while state-of-the-art (SOTA) methods may perform high-quality edits, they are likely to include unwanted changes of other facial features in incremental editing scenarios, where an already edited image is subject to further modifications using different modalities. Some existing methods~\cite{TediGAN_2021_CVPR,huang2022poegan} have the limitation in manual annotations of paired data. These methods train their models with labeled paired data across different modalities but manual annotations of training datasets are label-intensive. Recent methods based on diffusion models~\cite{huang2023collaborative,nair2023unite} instead train all uni-modal models first and perform multimodal facial editing by integrating these pre-trained uni-modal models. However, when the number of modalities grows, more uni-modal models should be trained separately with these methods. 

With this in mind, we explored ways to tackle these limitations. We investigated whether incorporating generative adversarial networks (GANs)~\cite{Goodfellow2014} can improve global consistency for multimodal local facial editing.
By learning the distribution of real facial images, adversarial training enforces the model to fill plausible contents for edited regions guided by multimodalities. 
To minimize the dependency on paired training data, we asked the question that if we can loosen the ties between the paired modalities data by aligning all modalities into a unified generative latent space to diminish the requirement for paired text, attribute label, and exemplar modalities. 
Instead of training a uni-modal model for each modality, we examined whether fusion and warping priors, along with multimodalities in both latent and feature space, could achieve seamless integration of multimodalities.

We thus introduce a  \textbf{MU}ltimodal \textbf{G}enerative and fusion framework for local \textbf{FAC}ial \textbf{E}diting (FACEMUG), which can solve the above problems. 
First, since the StyleGAN latent space~\cite{Karras2020} is disentangled well, we design our framework by bridging all modalities to the StyleGAN latent space.
Second, for the seamless integration of multimodalities, we design our multimodal generator with fusion and warping in latent and feature space.
To support the heterogeneity and sparsity of the pixel-wise conditional inputs, we aggregate multimodal conditional inputs into a homogeneous feature space.  
Since a fully trained GAN model excels at capturing rich textures and structural priors~\cite{chan2021glean},  we utilize a StyleGAN generator as a facial feature bank to provide candidate facial features and introduce style fusion blocks to fuse facial features for improving the generation quality. 
Moreover, to rectify the pose misalignment between the edited image and the given latent codes, we present a self-supervised latent warping method to efficiently transfer the pose of the edited image to that of the given latent codes in the latent space. To simulate the latent editing process during training and boost facial editing capabilities, a diversity-enhanced attribute loss is proposed.

 To the best of our knowledge, our FACEMUG is the first method that generates realistic facial features in response to multimodal inputs on the edited regions while maintaining visual coherence with the unedited background to achieve global consistency. 
 We have conducted extensive comparisons of FACEMUG against the SOTA methods and comprehensive experiments to demonstrate the superiority of FACEMUG in terms of editing quality, flexibility, and semantic control, illustrating its potential to significantly enhance various applications within facial editing. 

In summary, our paper makes the following contributions:
\begin{itemize}

     \item A novel globally-consistent local facial editing framework that enables diverse facial attribute manipulation.
	

    \item A novel multimodal feature fusion mechanism that utilizes multimodal aggregation and style fusion blocks to fuse facial features in both latent and feature spaces.


	\item A novel latent warping algorithm automatically aligns facial poses between edited and exemplar images in latent space, without relying on annotated labels or pose detection models.

	\item Our novel framework would benefit numerous practical applications, supporting incremental editing scenarios guided by multimodities (sketches, semantic maps, color maps, exemplar images, text, and attribute labels).  

\end{itemize}


\begin{figure*}[t]
	\centering
	\includegraphics[width=\textwidth]{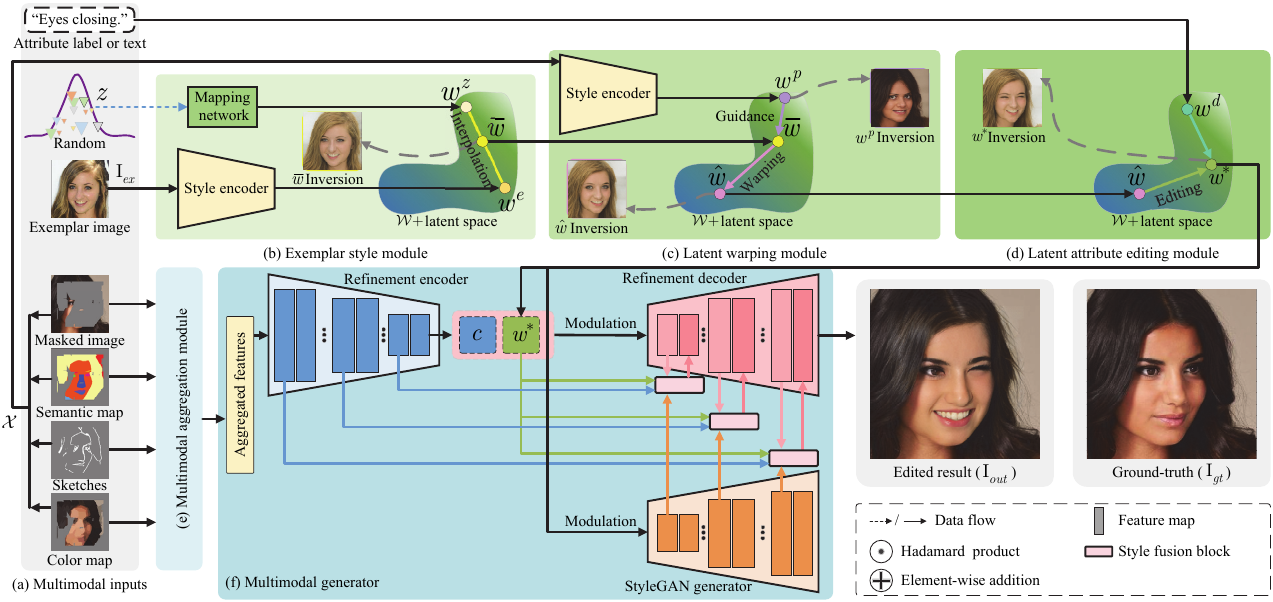}
	\caption{\revise{ Overall pipeline of our FACEMUG globally-consistent local facial editing: the given attribute label (or text), random latent code $z$, and exemplar image $\mathbf{I}_{ex}$ are first processed through the exemplar style module, latent warping module, and the latent attribute editing module to get the edited latent codes. Simultaneously, the input pixel-wise multimodal inputs $\mathcal{X}$ and a binary mask $\mathbf{M}$ are fed into the multimodal aggregation module and the multimodal generator to get an edited realistic face image $\mathbf{I}_{out}$, where the manipulation of the masked regions in $\mathbf{M}$ is guided by multimodal inputs. 
 }
	}\label{fig:architecture}
\end{figure*}


\section{Related work}\label{sec:relatedwork}

\subsection{Generative image synthesis}
GANs have been at the forefront of image synthesis in recent years credited to their ability to efficiently sample high-resolution images with good perceptual quality~\cite{Karras2020}.
The latent space learned by these networks enables an intuitive approach to controlling the image generation process, thereby facilitating semantic manipulation. Advancements in disentangled latent space learning, exemplified by developments in Fader Networks~\cite{NIPS2017_3fd60983} and StyleGANs~\cite{Karras2019,Karras2020,Karras2021}, have made the manipulation of latent codes in pre-trained GAN models an active area of research in image editing.

Another category of methods known as Diffusion Probabilistic Models~\cite{Nonequilibrium_dickstein15,NEURIPS2020DDPM} has demonstrated remarkable potential in enhancing image synthesis quality~\cite{dhariwal2021diffusion}. For instance, Latent Diffusion Models (LDMs, Stable Diffusion)~\cite{Rombach_2022_CVPR} were introduced to achieve better visual quality on image inpainting and class-conditional image synthesis, while significantly reducing computational requirements. Even though the sampling speed can be partially solved by sampling strategies~\cite{song2021denoising} and hierarchical approaches~\cite{Rombach_2022_CVPR,vahdat2021scorebased}, multiple denoising steps are still needed for high-quality synthesis. 
Compared to diffusion probabilistic models~\cite{Nonequilibrium_dickstein15,NEURIPS2020DDPM} with multiple denoising steps, FACEMUG ensures rapid inference without compromising the quality of the edited images.



\subsection{Facial attribute manipulation}~\label{sec:related_facialediting}
Facial attribute manipulation is a crucial task in computer graphics and computer vision, focusing on the modification and control of various facial features, such as skin texture~\cite{Jo2019}, structure~\cite{Wu_TOG2021}, and expressions~\cite{lu2022inpainting}. 
Existing image inpainting techniques~\cite{Yi2020,lu2023grig,Zhao2021} can directly remove unwanted or blemished facial features, replacing pixels in masked regions with plausible content. However, these techniques are limited in terms of controllable editing ability and do not support multimodal conditional editing.
Low-level facial features such as texture and geometry can be manipulated utilizing auxiliary information, such as sketches~\cite{Faceshop2018,Jo2019}, colors~\cite{Jo2019}, foreground contours~\cite{Xiong2019}, and structures~\cite{Ren2019}. However, these methods may overfit limited guidance information, requiring professional skills for semantic editing, such as facial expression and identity. 

For semantic-level face manipulation, some methods~\cite{Choi2018,Choi2020} can control a set of attributes using a domain label to index the mapped latent codes. However, they are limited to pre-defined attributes, thereby restricting editing freedom.  Geometry-guided face manipulation methods mainly use semantic geometry, such as sketches~\cite{Richardson2021,chenDeepFaceEditing2021,zhang2023adding}, semantic maps~\cite{Lee2020,ChenTOG2022,Zhu_2020_SEAN_CVPR,shi2021SemanticStyleGAN,wang2022semantic,IDE_3D_2022} to guide the generation of facial structure.
However, due to information loss during the projection and reconstruction process between real photographs and corresponding latent representations, these methods might inadvertently alter fine facial details (i.e., unedited regions may be changed).
Recent advancements allow the transfer of facial attributes from example images at the instance level~\cite{Xiao2018,Guo2019,Li2020_face,Liu_2021_CVPR,Chen2020_simSwap,Li2021}, manipulating attributes such as expression, identity, and decorative elements. 
However, these methods lack flexibility in selecting facial regions for local editing. 
Image-composition-based methods enable local editing by overlaying the foreground of a source image onto the background of a target image~\cite{zhao2019_guided_inpaint,Zhang2021_recomposition,kim2021stylemapgan}. Although they can generate more diverse and realistic facial images,  their method may fail when the poses in the edited and reference images differ.

Manipulating latent codes in pre-trained GAN models is another editing branch.
Studies have explored semantically meaningful paths for editing latent codes through supervised~\cite{Wu_TOG2021,Wu_2022_CVPR,shen2020interfacegan,Abdal_StyleFlow2021,CLIP2StyleGAN2022}  and unsupervised learning methods~\cite{Shen2020ClosedFormFO,harkonen2020ganspace}. With the advantages of StyleGAN's disentanglement in latent space, text descriptions employed by StyleCLIP~\cite{Patashnik_2021_ICCV}, FFCLIP~\cite{zhu2022one}, and CLIPInverter~\cite{CLIP_text_editing2023} are gaining popularity for text-based image manipulation. 
\revise{These methods relied on image inversion techniques}, might unintentionally alter unedited areas due to information loss in the inversion process, even though existing optimization-based~\cite{Abdal2019,Abdal_2020_cvpr,Pivotal_tuning2022} and encoder-based~\cite{Zhu2020,Richardson2021,Tov2021,wang2022HFGI,ReGANIE2023} GAN inversion methods can alleviate this issue.
In contrast, our method can manipulate local facial attributes with multimodal conditional information to produce realistic locally edited face images with fine-grained and semantic manipulation, preserving unaltered regions.

\subsection{Multimodal image synthesis}
Since different input modalities are well suited for manipulating  different aspects of facial attributes,
leading to recent studies exploring the multimodal image synthesis~\cite{Mixture_Shi_2019,sutter2021generalized,suzuki2017joint,zhan2023mise}.
Recently, TediGAN~\cite{TediGAN_2021_CVPR} leverages a pre-trained unconditional generator for high-fidelity multimodal facial editing. However, its approach of fusing multimodalities solely in the latent space limits its responsiveness to specific input modalities, particularly for detailed facial features.  
PoE-GAN~\cite{huang2022poegan} improves the visual quality and control abilities using a product-of-experts inference framework to learn the image distribution conditioned on multimodalities. However, it requires a specific feature extractor for each modality, which may not make good use of feature correlations between multimodalities. Moreover, this approach increases training parameters and complexity as more modalities are added. 
Collaborative-Diffusion~\cite{huang2023collaborative} and Unite\&Conquer~\cite{nair2023unite} demonstrate impressive multimodal generation and editing performances using pre-trained diffusion models. However, these methods necessitate multiple denoising steps for each uni-modal model during inference, resulting in slow inference speeds and significant GPU memory usage, especially when more modalities are involved in the editing process.
In this paper, our FACEMUG achieves seamless integration of multimodalities for globally consistent local facial editing and has the highest number of modalities.





\section{Method}\label{sec:algo}

\subsection{Overview}
The overall editing pipeline of our FACEMUG framework is shown in Fig.~\ref{fig:architecture}. Given a collection of  pixel-wise multimodal inputs $\mathcal{X}=\{\mathbf{I}_{m},\mathbf{I}_1,\mathbf{I}_2, \ldots,\mathbf{I}_n\}$, an attribute label ${w}^d \in \mathcal{W+}$ (or text $t_{tar}$), an exemplar image $\mathbf{I}_{ex}\in \mathbb{R}^{h \times w \times 3}$, a ground-truth face image $\mathbf{I}_{gt} \in \mathbb{R}^{h \times w \times 3}$ (with $h \times w$ pixels and three color channels), and a binary mask $\mathbf{M} \in \mathbb{R}^{h \times w \times 1}$ (with 1 for editing and 0 for unedited pixels), the masked image $\mathbf{I}_{m} \in \mathbb{R}^{h \times w \times 3}$ is obtained by $\mathbf{I}_{m} = \mathbf{I}_{gt} \odot (\mathbf{1}-\mathbf{M})$, where each input pixel-wise modality $\mathbf{I}_k \in \mathbb{R}^{h \times w \times c_k} $ contains $c_k$ channels and $\odot$ denotes the Hadamard product. Let $\mathcal{W}+$ denote the disentangled style latent space~\cite{Karras2020} and $\mathcal{Z}$ denote the random latent space. 
The goal of FACEMUG is to generate an edited realistic face image $\mathbf{I}_{out}$, where the manipulation of the masked regions in $\mathbf{M}$ is guided by multimodal inputs, while the unedited regions remain unchanged.  

\subsubsection{Exemplar style module} As shown in Fig.~\ref{fig:architecture} (b), our exemplar style module is designed to support randomized and exemplar-guided facial attribute editing. First, a multi-layer fully-connected neural mapping network $F_{\hat{\theta}_{f}}$ with the network parameters $\hat{\theta}_{f}$ linearly maps a random latent code  $z \in \mathbb{R}^{512 \times 1}$ ($z \in \mathcal{Z} $) to style latent codes ${w}^z = F_{\hat{\theta}_{f}}(z) = \left\lbrace w^z_{i} \in \mathbb{R}^{512 \times 1} | i \in T \right\rbrace \in \mathcal{W}+$, where $T = \lbrace1, 2,..., t\rbrace$ and $t$ is the number of the style latent code. Simultaneously, a style encoder $E_{\hat{\theta}_{e}}$ with the network parameters $\hat{\theta}_{e}$ maps multimodalities to  $\mathcal{W}+$. Given an exemplar image $\mathbf{I}_{ex}$, the style encoder extracts exemplar latent codes ${w}^e = \left\lbrace w^e_{i} \in \mathbb{R}^{512 \times 1} | i \in T \right\rbrace = E_{\hat{\theta}_{e}}(\mathbf{I}_{ex}) \in \mathcal{W}+ $.
Then we perform style interpolation between ${w}^z$ and ${w}^e$ to get the interpolated latent codes $\overline{w} \in \mathcal{W+}$.

\subsubsection{Latent warping module} As shown in Fig.~\ref{fig:architecture} (c), to alleviate visual artifacts due to the misalignment between the facial poses of the exemplar image and the edited image, we introduce a latent warping module $H_{\theta_{h}}$ that learns to transfer the pose of a target image to a source image in the style latent space with the learnable network parameters  $\theta_{h}$.  We first utilize the style encoder to project the pixel-wise multimodal inputs $\mathcal{X}$ to the projected latent codes ${w}^p = E_{\hat{\theta}_{e}}(\mathcal{X}) \in \mathcal{W}+$.
\revise{Then we obtain the warped latent codes as $\hat{w} =  H_{\theta_{h}}(w^{p} - \overline{w},\overline{w}) + \overline{w}$, $\hat{w} \in \mathcal{W}+$, under the guidance of $w^{p}$. 
As a consequence, $\hat{w}$ aligns to the pose of projected latent codes $w^{p}$ while preserving facial features of $\overline{w}$.}


\subsubsection{Latent attribute  editing module} 
As shown in Fig.~\ref{fig:architecture} (d), the latent attribute editing module is designed to edit the warped latent codes $\hat{w}$ to support attribute-conditional facial attribute editing and text-driven facial attribute editing in the style latent space. In this module, we obtain the edited latent codes ${w}^*$.

\subsubsection{Multimodal aggregation module} 
As shown in Fig.~\ref{fig:architecture} (e), for better controllability and visual quality in the image space, a multimodal aggregation module ${A}_{\theta_a}$ with the learnable network parameters $\theta_{a}$ is proposed to deal with multiple heterogeneous and sparse conditional inputs by merging them into a homogeneous feature space containing $c_{a}$ feature channels. Given the pixel-wise multimodal inputs $\mathcal{X}$, we obtain the aggregated feature tensor $\hat{\mathbf{F}}^{a} = {A}_{\theta_a}(\mathcal{X}) \in \mathbb{R}^{h \times w \times c_{a}}$. $\hat{\mathbf{F}}^{a}$ is then input into our multimodal generator.

\subsubsection{Multimodal generator}\label{sec:multi_generator}
As shown in Fig.~\ref{fig:architecture} (f), our multimodal generator consists of a facial feature bank and a refinement auto-encoder $G_{\theta_g}$ with the trainable network parameters $\theta_{g}$. We implement the facial feature bank using the StyleGAN generator $S_{\hat{\theta}_{s}}$. $S_{\hat{\theta}_{s}}$ produces multi-scale coarse facial feature maps from edited latent codes ${w}^*$ while $G_{\theta_g}$ refines the editing results by utilizing the aggregated feature tensor, the latent codes, and the generated coarse features. A set of facial feature maps $\mathcal{F}^s =  \{\mathbf{F}^s_i \in \mathbb{R}^{\hat{h}_{i} \times \hat{w}_{i} \times \hat{c}_{i}} | i \in T\}$ and a reconstructed image $\mathbf{I}_p$ are obtained as $(\mathcal{F}^s,\mathbf{I}_p) = S_{\hat{\theta}_{s}}({w}^*)$. We define $\hat{h}_i \times \hat{w}_i \times \hat{c}_i$ as the size of feature maps at $i$-th layer. Then, the refinement auto-encoder   $G_{\theta_g}$ leverages $\hat{\mathbf{F}}^{a}$, $w^*$, and $\mathcal{F}^s$ to generate an edited image $\mathbf{I}_{out} \in \mathbb{R}^{h \times w \times 3}$:
\begin{equation}\label{equ:refine_editing}
	\begin{aligned}
		\mathbf{I}_{out}(w^{*}) = \mathbf{I}_{m} \odot (\mathbf{1} - \mathbf{M}) + G_{\theta_g}(\hat{\mathbf{F}}^{a},w^{*},\mathcal{F}^s) \odot \mathbf{M}.
	\end{aligned}
\end{equation}
The refinement auto-encoder can be further divided into an encoder $G^{en}$ and a decoder $G^{de}$, i. e., $G_{\theta_g}= \{G^{en},G^{de}\}$.

\subsubsection{Discriminator} A discriminative network $D_{\theta_{d}}$ with the learnable network parameters $\theta_{d}$ learns to judge whether an image is a real or fake image. The discriminator maps an image (e.g., $\mathbf{I}_{out}$ or $\mathbf{I}_{gt}$) to a scalar $D_{\theta_{d}}(\mathbf{I}) \in \mathbb{R}^{1 \times 1}$. Note that the discriminator is only applied during the training phase.


\subsection{Latent warping module}\label{sec:latent_warping}
To achieve realistic exemplar-guided or randomized facial attribute editing, the primary challenge lies in the potential differences in pose between the exemplar image and the edited image, which can readily result in noticeable misalignment between the two facial images. Several existing methods~\cite{bounareli2022finding,Tewari2020StyleRigRS} have been developed for facial pose alignment in the context of face reenactment. 
However, these methods often require additional pre-trained facial pose detection models or involve multi-stage training processes, which can be resource-intensive. We thus ask: Is it possible to directly conduct facial pose transfer in the latent space, thereby eliminating auxiliary steps such as preliminary GAN inversion, subsequent pose detection, and ultimately pose transfer?

To solve this, we propose a self-supervised latent warping method that eliminates the need for manual annotations and pre-trained facial pose detection models. 
Our method provides an intuitive and straightforward way of warping the pose of the exemplar image to match that of the edited image in the latent space while preserving the attributes of individual faces (e.g., identity or expressions). The key idea is that we use the pose of a target image to guide the pose of a source image by warping the latent codes in the style latent space.

To effectively predict the offsets in the latent space from given two latent codes for facial warping, we design our latent warping network $H_{\theta_{h}}$ by four stacked code-to-code modulation blocks. 
We construct our code-to-code modulation block by extending FFCLIP's semantic modulation block~\cite{zhu2022one} to code-to-code embeddings. 
Moreover, the sigmoid activation is incorporated in the semantic injection for gate activation.

\revise{
Let $w^{ta} \in \mathbb{R}^{t \times 512 \times 1}$ and $w^{so} \in \mathbb{R}^{t \times 512 \times 1}$ be the target and source latent codes, respectively.
The warped latent codes $w^{wa} \in \mathbb{R}^{t \times 512 \times 1}$ is obtained by warping $w^{so}$ guided by $w^{ta}$:
\begin{equation}\label{equ:warping_qua}
		\begin{aligned}
            w^{wa} = H_{\theta_{h}}(w^r,w^{so})+w^{so}, 
            \end{aligned}
\end{equation}
where $w^{r}= w^{ta}-w^{so}$ is the residual latent codes between $w^{ta}$ and $w^{so}$. By leveraging the code-to-code modulation mechanism, our latent warping module effectively aligns the pose of the warped latent codes with the target latent codes.
}

\subsection{Latent attribute editing module} 

\revise{Our method supports two types of latent attribute editing, including attribute-conditional facial attribute editing and text-driven facial attribute editing. It allows us to utilize conditional labels or text to manipulate latent codes for semantic-level editing with unedited portions unchanged, which is usually hard to achieve with GAN-inversion-based methods~\cite{wang2022HFGI}.}

\textit{Attribute-conditional editing.} Each attribute label corresponds to a semantic direction. For a user-specified target attribute label, we obtain the edited latent codes $w^{*} \in  \mathcal{W+}$ by moving the warped latent codes $\hat{w}$ (${w}^{wa}$)  along the corresponding semantic direction $w^d$. The editing process can be expressed below~\cite{shen2020interpreting}:
\begin{equation}\label{equ:latent_editing1}
	\begin{aligned}
		{w}^{*} &= \hat{w} + \epsilon \cdot {w}^d,
	\end{aligned}
\end{equation}
where $\epsilon$ is a user-specified weight of the latent semantic direction (attribute label) ${w}^d \in \mathcal{W+}$ to control the degree of attribute adjustment.


\textit{Text-driven editing.} For a user-specified target attribute text $t_{tar}$, we leverage CLIP~\cite{CLIP_2021} to find  text-driven latent codes by solving the following latent codes optimization problem:
\begin{equation}\label{equ:latent_editing2}
	\begin{aligned}
       {w}^{*} &  =  \underset{{w} \in \mathcal{W}+}{\arg \min } \left( \lambda_{clip} \cdot \mathcal{L}_{clip} (t_{tar},t_{src},w,\hat{w})+ \lambda_{reg} \cdot \left\|w-\hat{w} \right\|_2 \right),
	\end{aligned}
\end{equation}
where $t_{src} = \textit{``face''}$; $\lambda_{clip}\in [0.1, 1.0]$ and $\lambda_{reg}$ are used to balance the directional CLIP loss term and the regularization term. By  default, we set $\lambda_{clip}=0.05$ and $\lambda_{reg}=0.08$. 
The directional CLIP loss $\mathcal{L}_{clip}$~\cite{shamshad2023clip2protect} is employed to align directions between the text-image pairs of the original and edited images in the CLIP space:
\begin{equation}\label{equ:latent_editing3}
	\begin{aligned}
       \mathcal{L}_{clip} (t_{tar},t_{src},w,\hat{w}) &= 1 -
       \cos\left(\Delta T, \Delta I \right),\\
       \Delta T &= E_T\left(t_{tar} \right)-E_T\left(t_{src}\right), \\
       \Delta I &= E_I\left(\mathbf{I}_{out}(w)\right)-E_I(\mathbf{I}_{out}(\hat{w})),
	\end{aligned}
\end{equation}
where $E_T$ and $E_I$ are the text and image encoders of the CLIP model, $\mathbf{I}_{out}(w)$ and $\mathbf{I}_{out}(\hat{w})$ are obtained using Eq.~\ref{equ:refine_editing}.


\subsection{Multimodal aggregation module} 
To integrate multimodalities within a unified framework, the varying density and value range among images, sketches, semantic maps, and color maps present challenges~\cite{Popovic_2023_WACV}. For pixel-wise multi-conditional image editing, the discrepancies in information content across modalities make it difficult to apply standard convolution layers to capture the diverse characteristics of each modality, leading to suboptimal generation quality. Moreover, the differing levels of details and densities in the inputs can impact the visual appearance and realism of the generated images. To mitigate this issue,  we introduce a multimodal aggregation module that efficiently aggregates the multimodal inputs. We achieve this by using separate convolution layers for each modality and incorporating a normalized adaptive weighting mechanism to merge extracted features into a homogeneous feature space.  As a result, the module provides more robust representations, making it well-suited for handling multi-conditional image editing tasks.

\revise{
For the pixel-wise multimodal inputs $\mathcal{X}$, we first employ a residual block to extract feature maps for each modality,  resulting in a feature set. 
Using this set, a shared residual block is utilized to compute the contribution scores for each spatial point across all pixel-wise modalities, producing a contribution score map for each modality. 
Each score map adaptively weights the importance of each modality in a pixel-wise fashion for the aggregation process.
Thus, we can get the aggregated feature $\hat{\mathbf{F}}^{a} = {A}_{\theta_a}(\mathcal{X}) \in \mathbb{R}^{h \times w \times c_{a}}$.
This adaptive weighting mechanism allows the model to assign higher importance to informative and detailed pixel-wise modalities while reducing the impact of less informative inputs.
}

\subsection{Multimodal generator}\label{sec:multi_style}

To support multimodal conditional editing and generate high-quality editing results, we develop a multimodal generator that fully utilizes aggregated facial features and edited latent codes, while efficiently fusing feature maps from the refinement encoder, the facial feature bank, and the refinement decoder. 
To achieve this, we introduce a style fusion block to fuse features in both high-level and shallow-level feature spaces. These carefully designed modules enhance our approach's editing capabilities, enabling the generation of diverse and high-fidelity editing results.

Given the  aggregated feature tensor $\hat{\mathbf{F}}^{a}$  and the edited latent codes $w^*$, the refinement encoder $G_{en}$ extracts multi-scale feature maps $\mathcal{F}^{en} = \{ \mathbf{F}^{en}_i \in \mathbb{R}^{\hat{h}_i \times \hat{w}_i \times \hat{c}_i}   | i \in T \}$ and outputs a global latent vector $c \in \mathbb{R}^{512 \times 2}$ from the aggregated multimodal feature $\hat{\mathbf{F}}^{a}$, i.e., $(\mathcal{F}^{en}, c) = G_{en}(\hat{\mathbf{F}}^{a})$. At the same time, the facial priors $\mathcal{F}^s =  \{\mathbf{F}^s_i \in \mathbb{R}^{\hat{h}_{i} \times \hat{w}_{i} \times \hat{c}_{i}} | i \in T\}$ are extracted from the edited latent codes $w^*$ in the facial feature bank with the StyleGAN generator. Finally, the feature maps $\mathcal{F}^{de}  = \{ \mathbf{F}^{de}_i \in \mathbb{R}^{\hat{h}_i \times \hat{w}_i \times \hat{c}_i}  | i \in T \} $ in the refinement decoder $G_{de}$ are calculated as follows: 
\begin{equation}\label{equ:decoder}
	\begin{aligned}
		\mathbf{F}^{de}_{i} = \begin{cases} \operatorname{UP}(\operatorname{SC}(\mathbf{F}^{de}_{i-1},[{w}^*_{i},c])), & \text {if } i \bmod 2 = 1;\\
			\operatorname{SC}(\mathbf{F}^{g}_{i-1},[{w}^*_{i},c]), & \text {otherwise}, \end{cases}
	\end{aligned}
\end{equation}
where $[\cdot,\cdot]$ denotes concatenation; $\operatorname{UP}(\cdot)$ refers to up-sampling; $\operatorname{SC}(\cdot,\cdot)$ indicates the style layer~\cite{Karras2020}, $\mathbf{F}^{de}_{0} = \operatorname{Conv}(\mathbf{F}^{en}_{t})$, $\operatorname{Conv}(\cdot)$ is a Convolution layer; $\mathbf{F}_{j}^{g}$ ($j = 1, 3, 5,\ldots, 2\lfloor \frac{t}{2} \rfloor+1$) is the fused feature map by the proposed style fusion block.
We set  $\mathbf{I}_{out} = \operatorname{Conv}(\mathbf{F}^{de}_{t}) \in \mathbb{R}^{h \times w \times 3} $ as the model output.

\begin{figure}[t]
	\centering
	\includegraphics[width=0.485\textwidth]{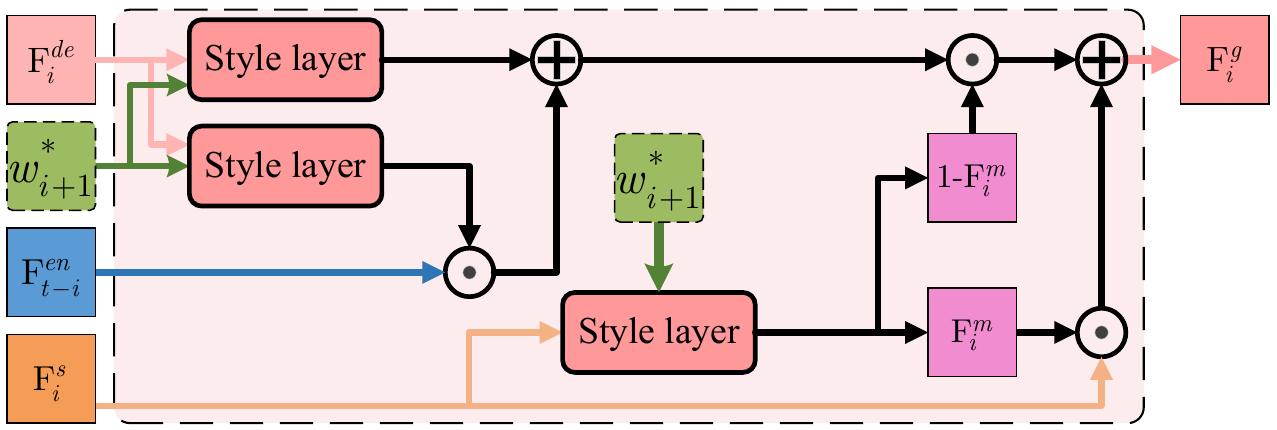}
	\caption{\revise{Illustration of our style fusion block. Conditioned by the modulated latent vector ${w}^*_{i+1}$, the block effectively integrates multi-scale facial features in both high-level and shallow-level feature spaces.}
	}\label{fig:fig_style_fusion_block}
\end{figure}

\textit{Style fusion block.}
The proposed style fusion block is as shown in \revise{Fig.~\ref{fig:fig_style_fusion_block}}. To fully leverage the guidance information extracted from the refinement auto-encoder and facial feature priors of the facial feature bank,  we employ the gated fusion scheme to perform element-wise fusion between these features for enhancement.
We first apply an adaptive gated fusion to activate features from the refinement encoder to obtain the intermediate generated feature $\hat{\mathbf{F}}^{g}_{i}$:
\begin{equation}\label{equ:gate_fusion}
		\begin{aligned}
			\hat{\mathbf{F}}^{g}_{i} = (\sigma(\operatorname{SC}(\mathbf{F}^{de}_{i},{w}^*_{i+1})) +&\mathbf{1})  \odot \mathbf{F}^{en}_{t-i} + \phi(\operatorname{SC}(\mathbf{F}^{de}_{i},{w}^*_{i+1})),
		\end{aligned}
\end{equation}
where $\sigma(\cdot)$ denotes the sigmoid activation function and $\phi(\cdot)$ corresponds to the LeakyReLU activation function with the negative slope of $0.2$. Then, the style layer computes the spatially-variant gate map $\mathbf{F}^{m}_i$ from the facial priors $\mathbf{F}^{s}_i$ and the modulated latent vector ${w}^*_i$ for each $i$-th layer. The feature fusion is calculated as follows:
\begin{equation}\label{equ:spatial_fusion}
		\begin{aligned}
			&\mathbf{F}^{m}_i =  \sigma(\operatorname{SC}(\mathbf{F}^{s}_i,{w}^*_{i+1})),\\
			&\mathbf{F}^{g}_i = \mathbf{F}^{m}_i  \odot \mathbf{F}^{s}_i + (\mathbf{1}-\mathbf{F}^m_i) \odot \hat{\mathbf{F}}^{g}_i,
		\end{aligned}
\end{equation}
where the spatially-variant gating map $\mathbf{F}^{m}_i$ automatically selects the important features from generated feature maps $\hat{\mathbf{F}}^{g}_i$ and the facial priors $\mathbf{F}^{s}_i$.

\begin{figure}[t]
	\centering
	\includegraphics[width=0.485\textwidth]{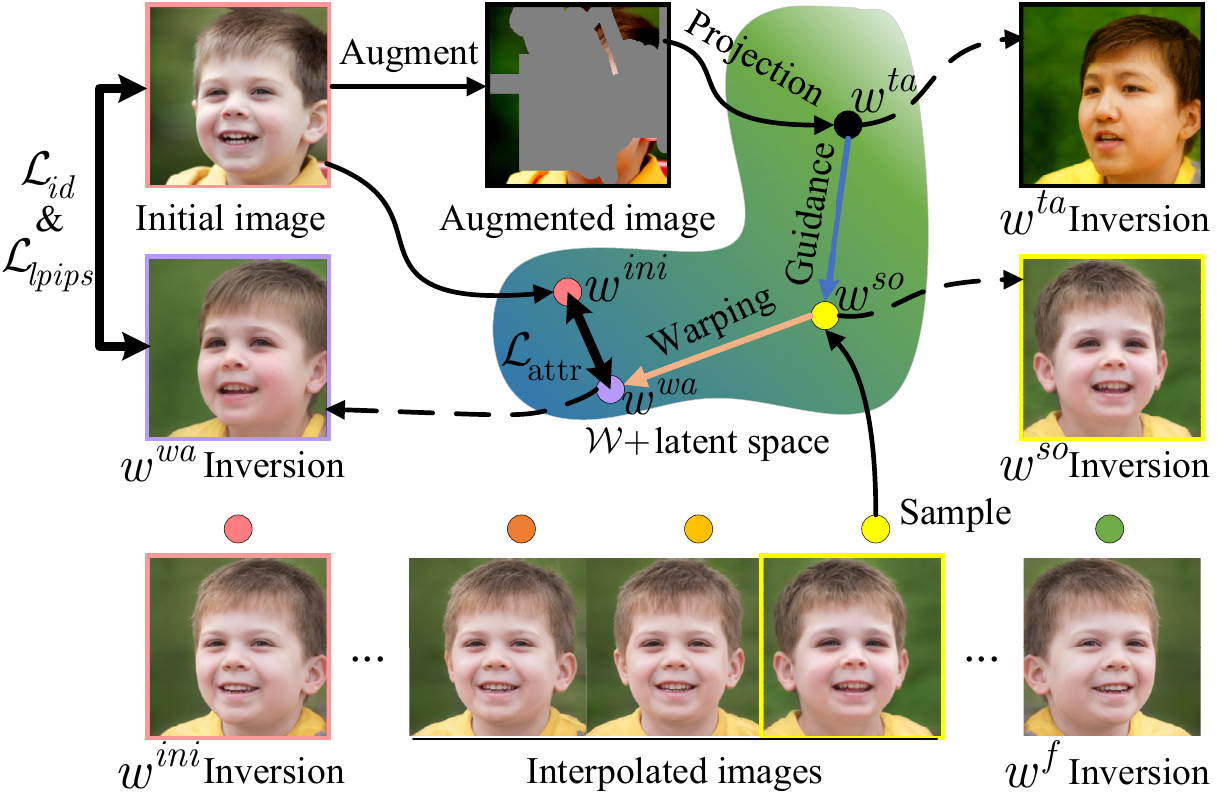}
	\caption{\revise{Illustration of the self-supervised training of our latent warping module. We employ the style encoder to project the augmented image to obtain the target latent codes $w^{ta}$. The source latent codes $w^{so}$ are sampled using interpolation between the initial latent codes $w^{ini}$ and the flipped latent codes $w^{f}$. The identity loss, the LPIPS loss, and the attribute loss are utilized as constraints to disentangle the identity and pose. This module effectively transfers the pose of $w^{ta}$ to the warped latent codes $w^{wa}$ while remaining other facial features unchanged. The inversion process is utilized for the visualization purpose.}}\label{fig:fig_latent_warping_train}
\end{figure}

\subsection{Self-supervised module training}\label{sec:training_FACEMUG}
In our framework, we utilize the pre-trained StyleGAN generator $S_{\hat{\theta}_{s}}$ and mapping network $F_{\hat{\theta}_{f}}$ from StyleGAN-V2~\cite{Karras2020}.
We begin by the training of the style encoder $E_{\hat{\theta}_{e}}$. Next, we optimize the latent warping network $H_{\theta_{h}}$ with our proposed self-supervised warping learning. Finally,  we detail the training process for the multimodal aggregation module ${A}_{\theta_a}$, the refinement auto-encoder $G_{\theta_g}$, and the discriminator $D_{\theta_{d}}$. This training process can be in parallel with the optimization of the latent warping network.


\subsubsection{Training of style encoder}  
\revise{The style encoder for exemplar images and that for multimodalities (sketch, color, semantic map, and mask) are the same encoder. The network of the style encoder $E_{\hat{\theta}_{e}}$ is borrowed from e4e~\cite{Tov2021}. 
 In order to enable the encoder to project each modality individually into the latent space, we customized the first convolution layer of the e4e encoder to handle 26 channels from four modalities respectively: 3 channels for the exemplar image, 1 channel for sketches, 3 channels for colors, and 19 channels for semantic layouts.
Then, we feed the concatenated randomly masked multimodal inputs into $E_{\hat{\theta}_{e}}$ for training. In addition, the loss functions defined in e4e~\cite{Richardson2021,Tov2021} are employed.}
\subsubsection{Training of latent warping module}
\revise{
As shown in Fig.~\ref{fig:fig_latent_warping_train}, a triplet of initial, source, and target latent codes is utilized to learn the pose warping guided by the target latent codes. We begin by projecting an initial image $\mathbf{I}_{ini}$ into the style latent space to get the initial latent codes $w^{ini} = E_{\hat{\theta}_{e}}(\mathbf{I}_{ini})$.
Then, we obtain an augmented image $\mathbf{I}_{ta}$ by applying bilinear scaling, color jittering, and region masking operations~\cite{Zhao2020_NIPS} to $\mathbf{I}_{ini}$. 
In addition, we obtain a flipped image $\mathbf{I}_f$ with mirror flipping of $\mathbf{I}_{ini}$. 
Therefore, $\mathbf{I}_f$ shares the same identity as $\mathbf{I}_{ini}$ with a flipped pose. Then, the target latent codes $w^{ta}$ and the flipped latent codes $w^f$ can be obtained by ${w}^{ta} = E_{\hat{\theta}_{e}}(\mathbf{I}_{ta})$ and ${w}^{f} = E_{\hat{\theta}_{e}}(\mathbf{I}_f)$, respectively. Next, we obtain the source latent codes ${w}^{so}$ with a linear interpolation:
\begin{equation}\label{equ:style_interpolation}
	\begin{aligned}
		{w}^{so}=  \beta  \cdot  w^{ini} + (1- \beta) \cdot {w}^f, \\
	\end{aligned}
\end{equation}
where $\beta \in [0,1]$ is a uniform random number. 
Since $w^{ini}$ and $w^{f}$ share the same facial features except for the pose, the interpolated source codes $w^{so}$ maintain the same facial identity to $w^{ini}$ but have a different pose. 
The warped latent codes ${w}^{wa}$ are obtained by transferring the pose of $w^{ta}$ to $w^{so}$ while maintaining the identity of $w^{so}$,
using our latent warping network $H_{\theta_{h}}$. 
Finally, we obtain the  warped image as $(\mathcal{F}^{wa},\mathbf{I}_{wa}) = S_{\hat{\theta}_{s}}({w}^{wa})$.
}




\revise{\textit{Total loss.} In order to disentangle the identity and pose during warping, we train $H_{\theta_{h}}$ by utilizing the identity loss, the LPIPS loss, and the attribute loss to constrain the identity and attribute similarities between $w^{wa}$ and $w^{ini}$.
The total training loss of $H_{\theta_{h}}$ is defined as:
\begin{equation}\label{equ:loss_total_latent}
	\begin{aligned}
			{O}(\theta_{h}) = \lambda_{latent} \cdot ({\mathcal{L}}_{id}(\mathbf{I}_{wa},\mathbf{I}_{ini}) + {\mathcal{L}}_{lpips}(\mathbf{I}_{wa},\mathbf{I}_{ini})\\+\mathcal{L}_{attr}({w}^{wa},w^{ini})),
	\end{aligned}
\end{equation}
where $\lambda_{latent}$ is empirically set to 0.1 in this work;
${\mathcal{L}}_{id}$, $\mathcal{L}_{lpips}$, and $\mathcal{L}_{attr}$ are the identity loss,
the LPIPS loss, and the attribute loss, respectively, and are defined below.}

\textit{Identity loss.} 
The identity  loss~\cite{lu2022inpainting,zhu2022one} is  incorporated to constrain the identity similarity:
\begin{equation}\label{equ:id_lpips_losses1}
	\begin{aligned}
		{\mathcal{L}}_{id}(\mathbf{I}_{x},\mathbf{I}_{y})= 1- \cos(R(\mathbf{I}_{x}),R(\mathbf{I}_{y})),
	\end{aligned}
\end{equation}
where $R(\cdot)$ is a pre-trained ArcFace network~\cite{Deng2019}. 

\revise{\textit{LPIPS loss.} 
The Learned Perceptual Image Patch Similarity (LPIPS)~\cite{Zhang2018} is applied to constrain the perceptual similarity:
\begin{equation}\label{equ:warp_lpips_losses2}
	\begin{aligned}
		{\mathcal{L}}_{lpips}(\mathbf{I}_{x},\mathbf{I}_{y})=\| P(\mathbf{I}_{x})-P(\mathbf{I}_{y})\|_{2},
	\end{aligned}
\end{equation}
where $P(\cdot)$ is a pre-trained VGG feature extractor~\cite{Simonyan2014}. }

\textit{Attribute loss.}
The attribute loss~\cite{lu2022inpainting,hou2022feat} is used to constrain the learning in the style latent space:
\begin{equation}\label{equ:loss_warping1}
	\begin{aligned}
		\mathcal{L}_{attr}({w}^{x}, w^{y})= \| {w}^{x}  - w^{y} \|_{2}.
	\end{aligned}
\end{equation}

\revise{Consequently, we can obtain the optimized parameters $\theta_{h}^*$ via the minimization of ${O}(\theta_{h})$. By taking advantage of the code-to-code modulation mechanism of $H_{\theta_{h}}$ with the loss constraints, ${w}^{wa}$ effectively learns the pose from $w^{ta}$ while remaining other facial features (e.g., identity) of $w^{so}$ unchanged.}


\subsubsection{Training of multimodal aggregation module, refinement auto-encoder and discriminator} 
The identity loss, the LPIPS loss, the diversity-enhanced attribute loss, and the adversarial loss are combined to optimize \revise{the multimodal aggregation module, the refinement auto-encoder, and the discriminator.

To learn the mapping between style latent codes and corresponding facial attributes, and to support attribute-conditional editing in the style latent space, a diversity-enhanced attribute loss $\mathcal{L}_{attr}({w}^o,\overline{w})$ is employed to constrain the consistency between facial attributes of the edited image $\mathbf{I}_{out}$ and the interpolated latent codes $\overline{w}$,
where ${w}^o= E_{\hat{\theta}_{e}}(\mathbf{I}_{out})$.

\textit{Total loss.}\label{sec:total_losses1}
The total training loss is defined as:
\begin{equation}\label{equ:loss_total1}
	\begin{aligned}
			{O}(\theta_{a},\theta_{g},&\theta_{d}) =  \lambda_{id}\mathcal{L}_{id}(\mathbf{I}_{out},\mathbf{I}_{ex})+\lambda_{attr}\mathcal{L}_{attr}({w}^o,\overline{w})\\
			&+ \lambda_{lpips}\mathcal{L}_{lpips}(\mathbf{I}_{out},\mathbf{I}_{gt})+\mathcal{L}_{adv}(\mathbf{I}_{out},\mathbf{I}_{gt}),
	\end{aligned}
\end{equation}
where we empirically set $\lambda_{id}=0.1$, $\lambda_{lpips}=0.5$, and $\lambda_{attr}=0.1$ in this work; $\mathcal{L}_{adv}$ is the adversarial non-saturating logistic loss~\cite{Goodfellow2014} with $R_{1}$ regularization~\cite{Mescheder2018}.

The refinement network $G_{\theta_g}$ is trained to generate a realistic edited image $\mathbf{I}_{out}$ while the discrinimator $D_{\theta_g}$ tries to  differentiate between $\mathbf{I}_{gt}$ and $\mathbf{I}_{out}$. 
In an alternating fashion, $A_{\theta_a}$ and $G_{\theta_g}$ are trained in a phase while $D_{\theta_d}$  is trained in the other. 
For each iteration, we obtain the optimized parameters ${\theta}^{*}_{a}$, ${\theta}^{*}_{g}$ and ${\theta}^{*}_{d}$ via the minimax game iteratively. 
}

\begin{figure*}[t]
	\centering
	\includegraphics[width=\textwidth]{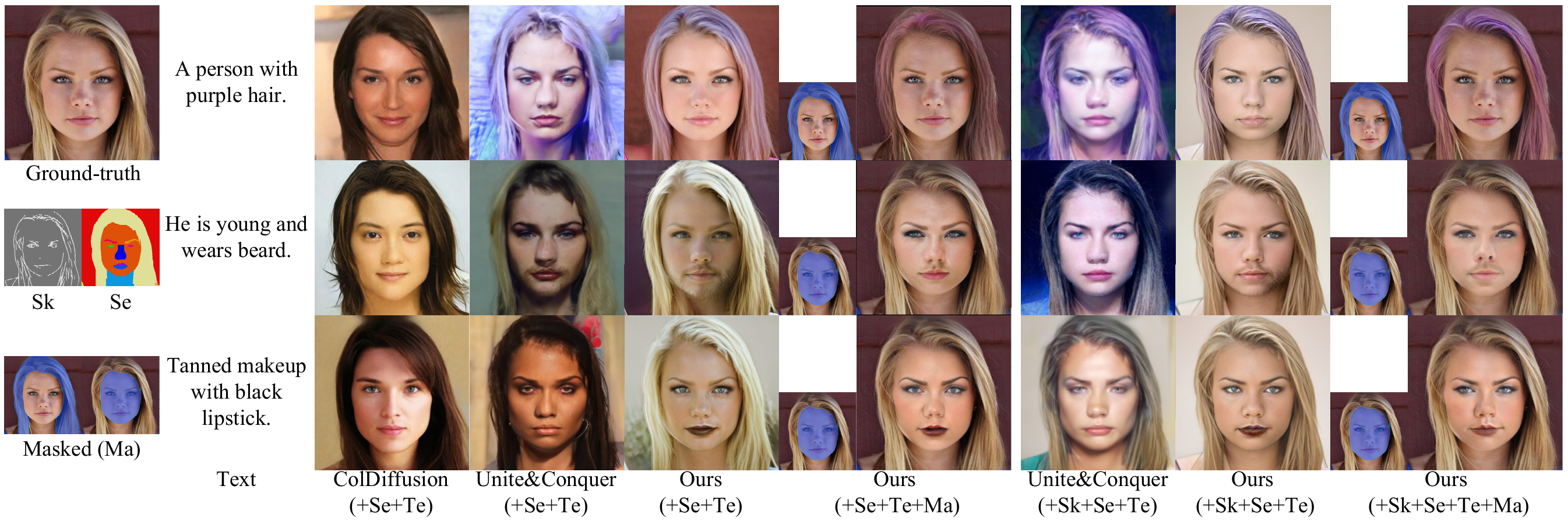}
	\caption{{Visual comparison to  ColDiffusion~\cite{huang2023collaborative} and Unite\&Conquer~\cite{nair2023unite} for text-driven multimodal facial editing. Our method produces visually appealing and globally consistent images with good responses to the corresponding multimodal inputs, and remains unmasked parts unchanged. }}\label{fig:fig_muli_CD_UQ_compare}  
\end{figure*}

\subsubsection{Implementation without manual annotation}\label{sec:implementation}

Our framework was implemented using Python and PyTorch. We trained FACEMUG and our latent warping module independently. 
\revise{For more implementation details of the latent warping module, the multimodal aggregation module, and the training procedures of our networks, please refer to the supplementary document.}

Manual annotations of labeled paired data across different modalities are required in existing multimodal editing methods~\cite{TediGAN_2021_CVPR,huang2022poegan}. On the contrary, the training modalities (i.e., editing mask, exemplar, semantics, sketch, and color) in this paper were generated without any manual annotation using well-built methods. 
The trained masks were generated randomly with the mask generation algorithm from CMOD~\cite{Zhao2021}. The Face-parsing model~\cite{face_parsing2023} was used to extract semantic maps. Hand-drawn-like sketches were generated using a pencil-sketch filter~\cite{Richardson2021}. \revise{Color images were processed using a mean color of each semantic region.} Exemplar images were sampled randomly from the ground-truth images. 
By aligning all modalities into a unified generative latent space, FACEMUG effectively loosens the ties between the paired modalities and enables model training without any human annotation.

The latent attribute editing module was implemented as follows. For the attribute-conditional editing, various attribute labels employed in InterfaceGAN~\cite{shen2020interfacegan}, GANSpace~\cite{harkonen2020ganspace}, StyleCLIP~\cite{Patashnik_2021_ICCV}, and CLIP2StyleGAN~\cite{CLIP2StyleGAN2022} were integrated in the module for semantic direction. For the text-driven editing, the edited latent codes were obtained through $100\sim300$ iterations of gradient descent~\cite{Patashnik_2021_ICCV} with the learning rate of $0.1$. 

Following the settings of StyleGANv2~\cite{Karras2020}, we employed the Adam optimizer with the first momentum coefficient of $0.5$, the second momentum coefficient of $0.99$, and the learning rate of $0.002$. 
We trained the networks for 800,000 iterations with a batch size of 8.

\section{Experimental results and comparisons}\label{sec:experiments}

\subsection{Settings}

\label{sec:exp_settings}

We conducted experimental evaluations on two publicly available and commonly used benchmark face image datasets: CelebA-HQ~\cite{Karras2018} and FFHQ~\cite{Karras2019}. 
Our FACEMUG was trained on the training set of FFHQ and evaluated on the testing set of  CelebA-HQ and FFHQ, respectively.  In the CelebA-HQ dataset, $2,000$ images were randomly selected for testing. For the FFHQ dataset, $60,000$ images were randomly chosen for training, and the remaining $10,000$ images were used for testing. 
All images were resized to the resolution of $256 \times 256$. To ensure a fair comparison, the same training and testing splits were used for all experiments. 

All experiments were conducted on the NVIDIA Tesla V100 GPU. The training time of our FACEMUG was around one month. We also evaluated FACEMUG on a PC equipped with an NVIDIA GeForce RTX 4090 GPU. It takes 29~ms (34~FPS) for each inference. 

\begin{figure*}[htbp]
	\centering
	\includegraphics[width=0.90\textwidth]{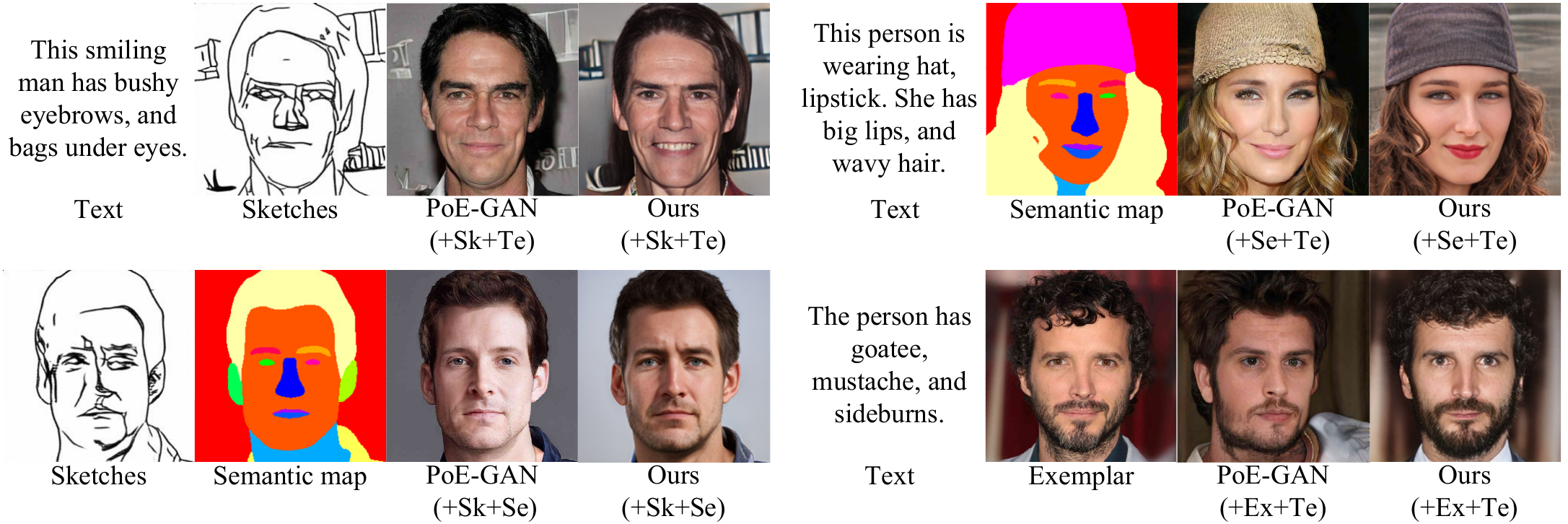}
	\caption{Visual comparison to PoE-GAN~\cite{huang2022poegan}. We used the input modalities and results published in their paper. The consistency between generated images and multimodal inputs of FACEMUG is better than that of PoE-GAN.
 }\label{fig:fig_muli_poe_gan_compare}  
\end{figure*}

We quantitatively evaluated the performance by using the  Fr\'{e}chet inception distance (FID)~\cite{Heusel2017}, the unpaired inception discriminative score (U-IDS)~\cite{Zhao2021}, and  LPIPS~\cite{Zhang2018} metrics which are robust assessment measures and correlate well with human perception for the image quality.

We demonstrated the performance of FACEMUG by utilizing various multimodal inputs. For convenience, we used the following abbreviations for these multimodalities: ``+Sk'' for adding sketches, ``+Se'' for adding semantic maps, ``+Co'' for adding color information,  “+Ex” for adding an exemplar image, “+Te” for adding text. \revise{We used the notation ``+Ma'' to represent the inclusion of a masked image for local editing. Different masks may affect the quantitative results because of the variation in position, size, and shape. We included various types of masks for comprehensive quantitative evaluation. 
Each mask was selected randomly from one of the following types of masks: hair, face, foreground subject, irregular region ($50-60\%$ mask ratio), and a fixed center ($128 \times 128$) rectangle. 
The center mask was included because it effectively covers most of the facial region in a facial image.
For quantitative comparisons, masks were obtained automatically.
Semantics-based masks (hair, face, etc.) were created with the Face-parsing model~\cite{face_parsing2023}. 
Irregular masks were obtained from the irregular mask templates~\cite{Liu2018}.
For qualitative comparisons, masks were manually defined.
A consistent set of inputs was used in each comparison to ensure fairness, both quantitatively and qualitatively.
}
For more experimental results, please refer to the supplementary document.



\subsection{Comparison on multimodal facial editing}\label{sec:comparison_Multimodal}
To evaluate the quality of generated images and the responsiveness of multimodal inputs, we performed comparisons against SOTA multimodal facial image editing techniques, including Unite\&Conquer~\cite{nair2023unite}, Collaborative-Diffusion (ColDiffusion)~\cite{huang2023collaborative}, and PoE-GAN~\cite{huang2022poegan} using multimodal conditional inputs. 
Unless specified, we employed officially released pre-trained models of compared methods.

\begin{figure*}[htbp]
	\centering
	\includegraphics[width=\textwidth]{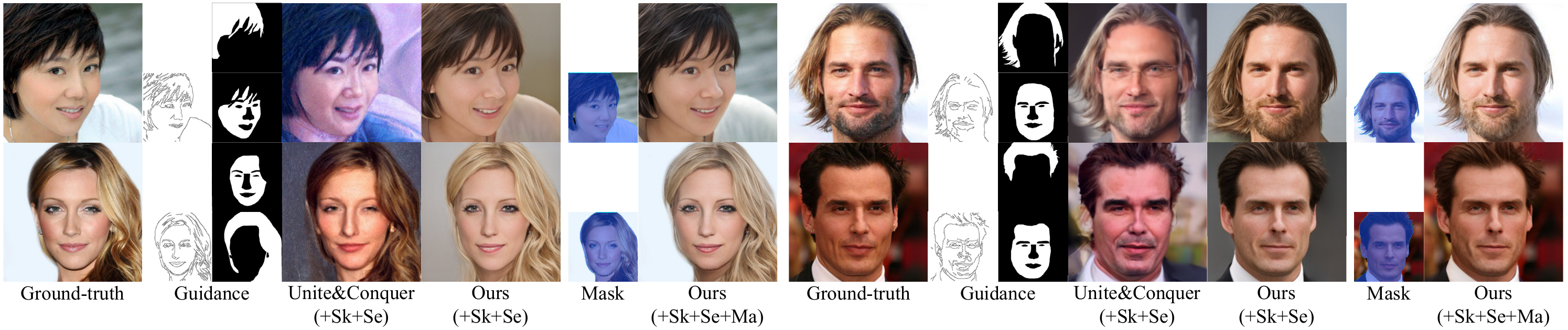}
	\caption{Visual comparison to Unite\&Conquer~\cite{nair2023unite}. We show more FACEMUG results by adding extra masks. Our method shows better visual quality and preserves background information when using masks.  }\label{fig:fig_multi_editing_qualitative}  
\end{figure*}

\begin{table}[t]
	\caption{Quantitative comparison of our method to SOTA multimodal facial editing methods with sketches, semantic maps, and text on the CelebA-HQ dataset. \textbf{Bold}: top-1 quantity.}
	\label{tab:mul_editing2}
	\centering
		\begin{tabular}{cccc}
			\toprule
			\makebox[0.5\columnwidth][c]{Method} & FID$^{\downarrow}$ & U-IDS$^{\uparrow}$& LPIPS$^{\downarrow}$  \\
			\midrule
			ColDiffusion~\cite{huang2023collaborative} (+Se+Te)    & 26.87  & 0 & 0.5283  \\
   			Unite\&Conquer~\cite{nair2023unite} (+Se+Te)   &   44.52  &  0  &   0.5809   \\
            Ours (+Se+Te)     & 38.41   & 0 & 0.4699  \\
            \revise{Ours (+Se+Te+Ma)}   & \revise{\textbf{11.85}}  & \revise{\textbf{0.20\%}} & \revise{\textbf{0.1645}} \\
            \midrule
			Unite\&Conquer~\cite{nair2023unite} (+Sk+Se+Te)   &  45.29  &  0  &  0.5493  \\
   			Ours (+Sk+Se+Te)     & 32.65    & 0 &  0.4046 \\
         	\revise{Ours (+Sk+Se+Te+Ma)}   & \revise{\textbf{11.24}}  & \revise{\textbf{0.28\%}} & \revise{\textbf{0.1448}} \\
			\bottomrule
	\end{tabular}
\end{table}

\begin{figure*}[t]
	\includegraphics[width=\textwidth]{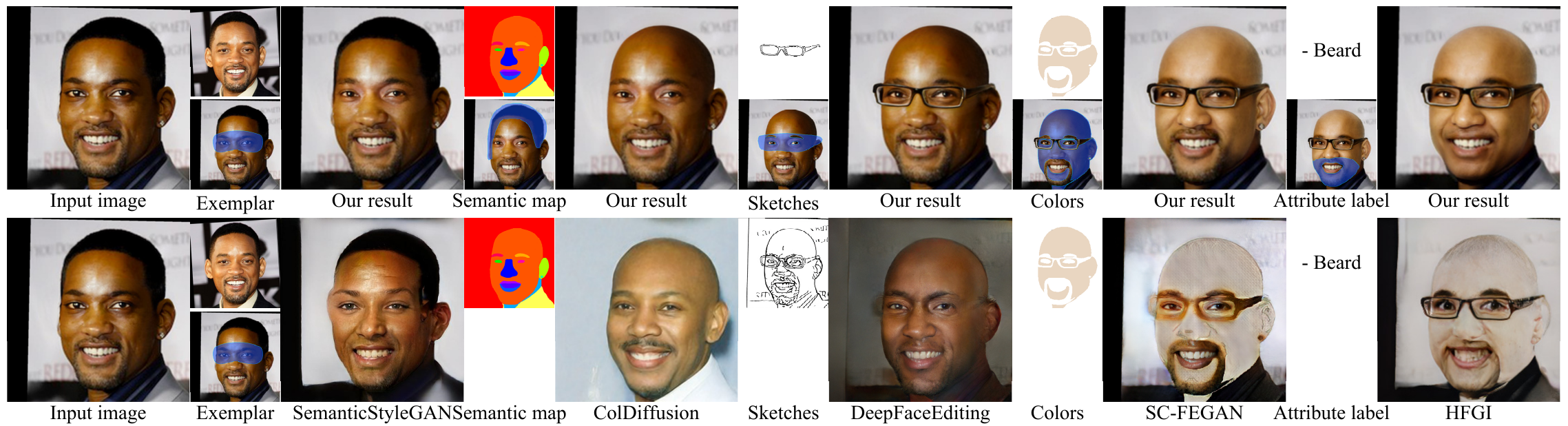}
	\caption{Qualitative comparison of incremental editing (the later editing taking the previous output image as input), compared to SOTA methods (SemanticStyleGAN~\cite{shi2021SemanticStyleGAN}, ColDiffusion~\cite{huang2023collaborative}, DeepFaceEditing~\cite{chenDeepFaceEditing2021}, SC-FEGAN~\cite{Jo2019}, HFGI~\cite{wang2022HFGI}) in row 2, using uni-modality. Our model shows better visual quality.}
	\label{fig:fig_incri_multi_comp}
\end{figure*}

Fig.~\ref{fig:fig_muli_CD_UQ_compare} shows visual comparisons of our FACEMUG to Unite\&Conquer and ColDiffusion. It shows that all the compared methods can use semantics or sketches to control the facial layout while adjusting appearance using given text. However, Unite\&Conquer and ColDiffusion show low consistency between the output image and the text caption. In contrast, our method is capable of combining the three modalities to perform high-quality multimodal editing. Fig.~\ref{fig:fig_muli_poe_gan_compare} shows visual comparisons of our FACEMUG  to PoE-GAN using text with sketches, semantics, and exemplar, respectively. The results of PoE-GAN show high-quality and good responses to the corresponding input modalities. However, for the first case, our method shows a clear smile expression, and for the last case, the edited result exhibits more correlation with the exemplar. For FACEMUG, the semantic maps and sketches provide geometry information, while text controls the appearance of the generated content.  Moreover, by leveraging the additional masked image, the generated content exhibits high consistency to the input modalities and is coherent to the editing boundary, preserving the unedited part unchanged.
We also show the quantitative comparison of multimodal editing with ColDiffusion~\cite{huang2023collaborative} and Unite\&Conquer~\cite{nair2023unite} on CelebA-HQ dataset, as shown in Table~\ref{tab:mul_editing2}. The semantic maps and text are from CelebAMask-HQ~\cite{Lee2020} and CelebA-Dialog~\cite{Jiang_2021_ICCV} datasets, respectively. 
Guided by semantic maps and text, both ColDiffusion and Unite\&Conquer produce high-quality images yet exhibit lower consistency with ground-truth images, as reflected by their LPIPS scores. 
In contrast, FACEMUG achieves lower LPIPS and competitive FID scores, indicating good fidelity of our results. 
When applying a mask to indicate the editing regions, FACEMUG not only demonstrates improved performance of FID and LPIPS scores but also illustrates superior editing quality. 
\revise{Our method achieves the lowest FID scores because it can modify specific facial attributes within the masked area while maintaining the unmasked regions unchanged and enhancing the overall image consistency.}


\begin{figure*}[t]
	\centering
	\includegraphics[width=\textwidth]{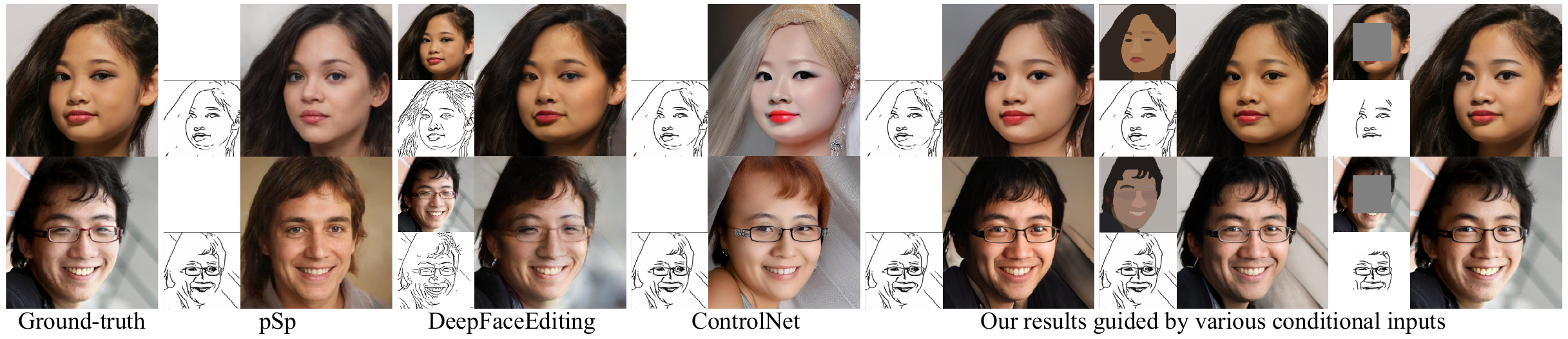}
	\caption{Visual comparison between our FACEMUG and the SOTA  sketch-guided editing methods (pSp~\cite{Richardson2021}, DeepFaceEditing~\cite{chenDeepFaceEditing2021}, and ControlNet~\cite{zhang2023adding}).  The sub-images in each group represent the guidance information for the editing process. FACEMUG produces images with superior quality and finer details by using more modalities, and shows global consistency when adding masks for local facial editing.  }\label{fig:fig_sketch_generatioin_qualitative}  
\end{figure*}

\begin{table}[t]
	\caption{Quantitative comparison to SOTA multimodal facial image editing methods with sketches and semantic maps on CelebA-HQ and FFHQ datasets. \textbf{Bold}: top-1 quantity.}
	\label{tab:multi_editing}
	\centering
	 \resizebox{\linewidth}{!}{
		\begin{tabular}{ccccccc}
			\toprule
			\multicolumn{1}{c}{\multirow{2}{*}{Method}} & \multicolumn{3}{c}{CelebA-HQ} &  \multicolumn{3}{c}{FFHQ} \\
			\cline{2-7}    \multicolumn{1}{c}{} & \multicolumn{1}{c}{FID$^{\downarrow}$} & \multicolumn{1}{c}{U-IDS$^{\uparrow}$}& \multicolumn{1}{c}{LPIPS$^{\downarrow}$} &  \multicolumn{1}{c}{FID$^{\downarrow}$} & \multicolumn{1}{c}{U-IDS$^{\uparrow}$}& \multicolumn{1}{c}{LPIPS$^{\downarrow}$} \\
			\midrule
			Unite\&Conquer~\cite{nair2023unite} &   \multirow{2}{*}{44.76}  &  \multirow{2}{*}{0}   & \multirow{2}{*}{0.5350}   &  \multirow{2}{*}{52.39}    & \multirow{2}{*}{0} &   \multirow{2}{*}{0.5707}  \\
            (+Sk+Se)  &    &     &    &     &  &     \\
			Ours (+Sk+Se)&   {29.96} &  0  & 0.3999  &  23.16 &  0.03\% &   0.4036  \\
                \revise{Ours (+Sk+Se+Ma)}&   \revise{\textbf{10.36}}  &  \revise{\textbf{0.63\%}}   &   \revise{\textbf{0.1371}}    & \revise{\textbf{2.26}} &  \revise{\textbf{30.38\%}} &  \revise{\textbf{0.1162}}  \\
			\bottomrule
		\end{tabular}
	 }
\end{table}

We further conducted comparisons between FACEMUG and Unite\&Conquer~\cite{nair2023unite}, focusing on facial editing using sketches and semantics. 
For a fair comparison, given that Unite\&Conquer utilizes only the  ``skin'' and ``hair''  semantic maps, we accordingly adjusted FACEMUG by removing other semantic labels.
As shown in Fig.~\ref{fig:fig_multi_editing_qualitative}, Unite\&Conquer unites multiple diffusion models trained on multiple sub-tasks to perform editing with the guidance of sketch and partial semantic labels. However, the presence of visual artifacts in the edited images indicates that employing disparate off-the-shelf diffusion models, trained on different datasets, remains a challenging task. In comparison, FACEMUG adeptly leverages sketches and partial semantic labels to facilitate geometry-guided facial generation (+Sk+Se) and editing (+Sk+Se+Ma). Moreover, the generated contents maintains consistency with the unedited portions.
Table~\ref{tab:multi_editing} displays the FID, U-IDS, and LPIPS scores of each method on CelebA-HQ and FFHQ datasets. 
Unite\&Conquer introduces a novel reliability parameter to facilitate the multimodal mixing of contents generated from various uni-modal diffusion networks. Nevertheless, our method surpasses Unite\&Conquer, exhibiting superior FID, U-IDS, and LPIPS scores.

\begin{figure*}[htbp]
	\centering
	\includegraphics[width=\textwidth]{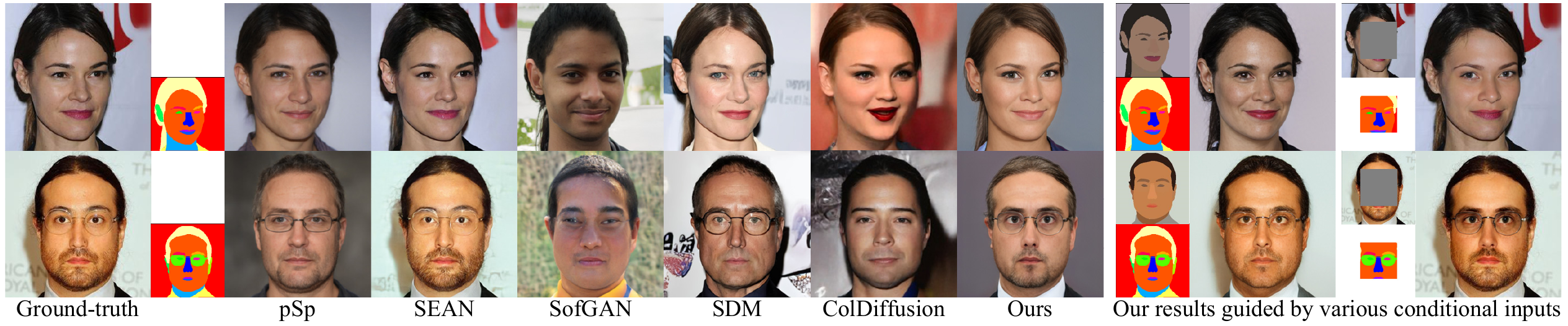}
	\caption{Visual comparison between our FACEMUG and the SOTA  semantic-guided editing methods (pSp~\cite{Richardson2021}, SEAN~\cite{Zhu_2020_SEAN_CVPR}, SofGAN~\cite{ChenTOG2022}, SDM~\cite{wang2022semantic}, and ColDiffusion~\cite{huang2023collaborative}). The sub-images in each group represent the guidance information for the editing process. FACEMUG produces more visually appealing results using more conditional modalities, and achieves high-quality local facial editing by incorporating masks. }\label{fig:fig_semantic_generatioin_qualitative}  
\end{figure*}

Fig.~\ref{fig:fig_teaser_multi} and Fig.~\ref{fig:fig_incri_multi_comp} show examples of incremental multimodal local facial editing. Our FACEMUG incrementally edits the input facial images to achieve high-quality manipulation by taking advantage of multimodal inputs, including masks, exemplars, semantics, sketches, colors, and attribute labels. 
It is worth noting that incremental editing is achieved effectively with our unified FACEMUG model, while existing methods have to use multiple uni-modal models (see Fig.~\ref{fig:fig_incri_multi_comp}) and introduce unwanted modifications on unedited facial features (see Fig.~\ref{fig:fig_teaser_multi}).  
By leveraging provided multimodalities, our FACEMUG can edit various realistic facial attributes (e.g., facial geometry, hairstyle, and decorative goods) while preserving unedited parts unchanged. 




\begin{table}[t]
	\caption{\revise{The user study results on the CelebA-HQ dataset. We present the percentages (\%) of cases where our results were preferred over those of the compared methods.} }
	\label{tab:user_study}
 	\centering
  {
        \revise{
		\begin{tabular}{ccc}
			\toprule
		Configuration &	\makebox[0.53\columnwidth][c]{Method-to-method comparison}&  Percentage  \\
			\midrule
		+Se+Te&	Ours vs. ColDiffusion~\cite{huang2023collaborative}    & 54.90\%  \\
   	+Se+Te&		Ours vs. Unite\&Conquer~\cite{nair2023unite} &    76.83\%  \\
        +Se+Sk+Te&    Ours  vs. Unite\&Conquer~\cite{nair2023unite} &  77.88\% \\  
		\bottomrule
	\end{tabular}
        }
        }
\end{table}


\revise{
\subsubsection*{User study} We conducted a user study for multimodal facial editing comparing ColDiffusion and Unite\&Conquer with  ``Se+Te'' and ``+Sk+Se+Te'' configurations.  
We sampled 100 images randomly from the CelebA-HQ test set and obtained corresponding edited images.
20 pairs of edited images were chosen randomly for each method-to-method comparison. For FACEMUG, ten edited images were produced with additional masks (+Ma), which were selected randomly from hair, face, and foreground subject masks; the other ten images were produced without masks.
The participants were asked to perform two-alternative forced choices (2AFCs) based on visual realism and consistency with input modalities.
Finally, we recruited 52 participants, resulting in 1040 votes per comparison.


Table~\ref{tab:user_study} shows the user study results.
With the ``+Se+Te'' configuration, our FACEMUG received $54.90\%$ and $76.83\%$ of the preference votes compared to ColDiffusion ($45.10\%$) and Unite\&Conquer ($23.17\%$), respectively.
Our FACEMUG also surpassed Unite\&Conquer with $77.88\%$ of majority votes with the ``+Sk+Se+Te'' configuration.
The user study validated that our FACEMUG effectively produces realistic facial images by taking advantage of multimodal local facial editing.
}

\begin{table}[!t]
	\caption{Quantitative comparison to SOTA sketch-guided facial image editing methods (+Sk) on CelebA-HQ and FFHQ datasets. \textbf{Bold}: top-1 quantity.}
	\label{tab:sketch_generation}
	\centering
	 \resizebox{\linewidth}{!}{
		\begin{tabular}{ccccccc}
			\toprule
			\multicolumn{1}{c}{\multirow{2}{*}{Method}} & \multicolumn{3}{c}{CelebA-HQ} &  \multicolumn{3}{c}{FFHQ} \\
			\cline{2-7}    \multicolumn{1}{c}{} & \multicolumn{1}{c}{FID$^{\downarrow}$} & \multicolumn{1}{c}{U-IDS$^{\uparrow}$}& \multicolumn{1}{c}{LPIPS$^{\downarrow}$} &  \multicolumn{1}{c}{FID$^{\downarrow}$} & \multicolumn{1}{c}{U-IDS$^{\uparrow}$}& \multicolumn{1}{c}{LPIPS$^{\downarrow}$} \\
			\midrule
			TediGAN~\cite{TediGAN_2021_CVPR} & 42.83   &  0   &  0.4909  & 85.17  &  0  &   0.6031   \\
			pSp~\cite{Richardson2021} &     42.70   &  0   & 0.4911 &  85.16 & 0  & 0.6031  \\
			DeepFaceEditing~\cite{chenDeepFaceEditing2021} &    19.78      &   0  &  0.2796  & 11.59 & 5.13\%  & 0.2850 \\
			ControlNet~\cite{zhang2023adding} &   64.59  &  0  &  0.5530  &   62.31  &  0  &   0.5475   \\
			Ours (+Sk)&  36.58   &  0   &   0.4071   & 24.29 &  0 &   0.4096\\
			Ours (+Sk+Co)&  17.43   & 0    &  0.2936    &  {9.12}  &  {9.59\%} &  0.2815 \\
                \revise{Ours (+Sk+Ma)}&  \revise{\textbf{11.43}}   & \revise{\textbf{0.52\%}}   &  \revise{\textbf{0.1444}}   &  \revise{\textbf{2.49}}  &  \revise{\textbf{29.51\%}} &   \revise{\textbf{0.1259}}  \\
			\bottomrule
	\end{tabular}
  }
\end{table}

\subsection{Comparison on sketch-guided facial editing}\label{sec:comparison_sketch}

We compared FACEMUG to the SOTA sketch-guided facial editing methods, including pSp~\cite{Richardson2021}, DeepFaceEditing~\cite{chenDeepFaceEditing2021} and ControlNet~\cite{zhang2023adding}.
The pre-trained models of pSp and DeepFaceEditing provided in the official online repository were used in this experiment. For ControlNet, we fine-tuned the officially provided pre-trained weights on the FFHQ training set with corresponding sketches. 
To ensure optimal performance, we used default forms of sketches that were used during training for each method.


As depicted in Fig.~\ref{fig:fig_sketch_generatioin_qualitative}, all the compared methods are capable of generating high-quality images guided by sketches. However, pSp may neglect some facial attributes, like glasses from input sketches.  Even conditioned on facial appearance and sketches, DeepFaceEditing exhibits some visual artifacts in edited outputs. While ControlNet can generate high-quality results, it needs several denoising steps and depends on high-quality prompts.
Without colors and unmasked background information (third last column), the color richness generated by FACEMUG may be affected. By incorporating both colors and sketches as guidance (second last column), FACEMUG can produce facial images with high fidelity. Our method (last column) demonstrates sketch-guided local facial editing, showcasing our method's ability to achieve high-quality facial image editing by leveraging unmasked pixels and input sketches while preserving the unedited regions unchanged.


Table~\ref{tab:sketch_generation} presents a quantitative comparison between FACEMUG and the existing SOTA approaches. All methods demonstrate good FID scores on both datasets. DeepFaceEditing achieves competitive FID and U-IDS scores by utilizing sketch and appearance information extracted from ground-truth images. Since sketches lack color and appearance information, the performance of FACEMUG (+Sk) may be limited. FACEMUG (+Sk+Co), which incorporates color as guidance information, further enhances the visual quality. FACEMUG (+Sk+Ma) surpasses all the compared methods by leveraging background information, demonstrating superior performance.

\subsection{Comparison on semantic-guided facial editing}\label{sec:comparison_semantic}
We compared FACEMUG to the SOTA semantic-guided facial editing methods, including 
pSp~\cite{Richardson2021}, SEAN~\cite{Zhu_2020_SEAN_CVPR}, SofGAN~\cite{ChenTOG2022}, SDM~\cite{wang2022semantic}, and ColDiffusion~\cite{huang2023collaborative}.
The pre-trained models of the compared methods provided in the official online repository were used in this experiment.


\begin{table}[t]
	\caption{Quantitative comparison to SOTA semantic-guided facial image editing methods (+Se) on CelebA-HQ and FFHQ datasets. \textbf{Bold}: top-1 quantity.}
	\label{tab:semantic_generation}
	\centering
	 \resizebox{\linewidth}{!}{
		\begin{tabular}{ccccccc}
			\toprule
			\multicolumn{1}{c}{\multirow{2}{*}{Method}} & \multicolumn{3}{c}{CelebA-HQ} &  \multicolumn{3}{c}{FFHQ} \\
			\cline{2-7}    \multicolumn{1}{c}{} & \multicolumn{1}{c}{FID$^{\downarrow}$} & \multicolumn{1}{c}{U-IDS$^{\uparrow}$}& \multicolumn{1}{c}{LPIPS$^{\downarrow}$} &  \multicolumn{1}{c}{FID$^{\downarrow}$} & \multicolumn{1}{c}{U-IDS$^{\uparrow}$}& \multicolumn{1}{c}{LPIPS$^{\downarrow}$} \\
			\midrule
			TediGAN~\cite{TediGAN_2021_CVPR} & 45.19 &  0   &  0.5208  &  85.50 & 0   &  0.6109   \\
			pSp~\cite{Richardson2021} &   45.19  &  0    &  0.5208   & 85.50   & 0 &  0.6109 \\
			SEAN~\cite{Zhu_2020_SEAN_CVPR} &   28.74   &  0   &  0.2595  &  26.55 &  1.5\%   &  0.3801  \\
			SofGAN~\cite{ChenTOG2022} &  38.88   &  0  & 0.6369  &  15.43 &  8.14\%  &  0.6297  \\
			SDM~\cite{wang2022semantic} &  27.61   &  0   &  0.4995  &  - &  -   &  -  \\
            ColDiffusion~\cite{huang2023collaborative} &  30.22   &  0   &  0.5280  &  76.73 &  0  &  0.5937  \\
			Ours (+Se)&  41.18  &   0  &   0.4462  & 43.78 &  0 &   0.4536 \\
			Ours (+Se+Co)& 31.02   &  0  & 0.3691  &  {14.08} &  {8.09\%} &   0.3018  \\
            \revise{Ours (+Se+Ma)}&  \revise{\textbf{10.65}} &  \revise{\textbf{0.53\%}}   &  \revise{\textbf{0.1570}}   & \revise{\textbf{ 2.41}} &  \revise{\textbf{28.99\%}} &  \revise{\textbf{0.1356}}  \\
			\bottomrule
		\end{tabular}
	 }
\end{table}

As shown in Fig.~\ref{fig:fig_semantic_generatioin_qualitative}, all methods can generate high-quality images guided by semantic maps. However, pSp and ColDiffusion may not accurately capture certain attributes (e.g., the shape of the glasses). SofGAN may produce facial images with less defined details. 
SDM and ColDiffusion are capable of generating superior images but at the cost of longer computation time.
Additionally, above compared methods fail to preserve the unedited regions surrounding the editing area. In contrast, FACEMUG (third last column) conditioned by semantic maps exhibits clear manifestations of semantic maps. When guided by color information as well (second last column), our method demonstrates superior visual performance. Furthermore, FACEMUG (last column) allows for local facial editing guided by semantic maps, preserving known regions while providing more flexibility for interactive facial attribute manipulation. FACEMUG consistently produces high-quality results with facial attributes guided by semantic maps.

\begin{figure}[t]
	\centering
	\includegraphics[width=0.40\textwidth]{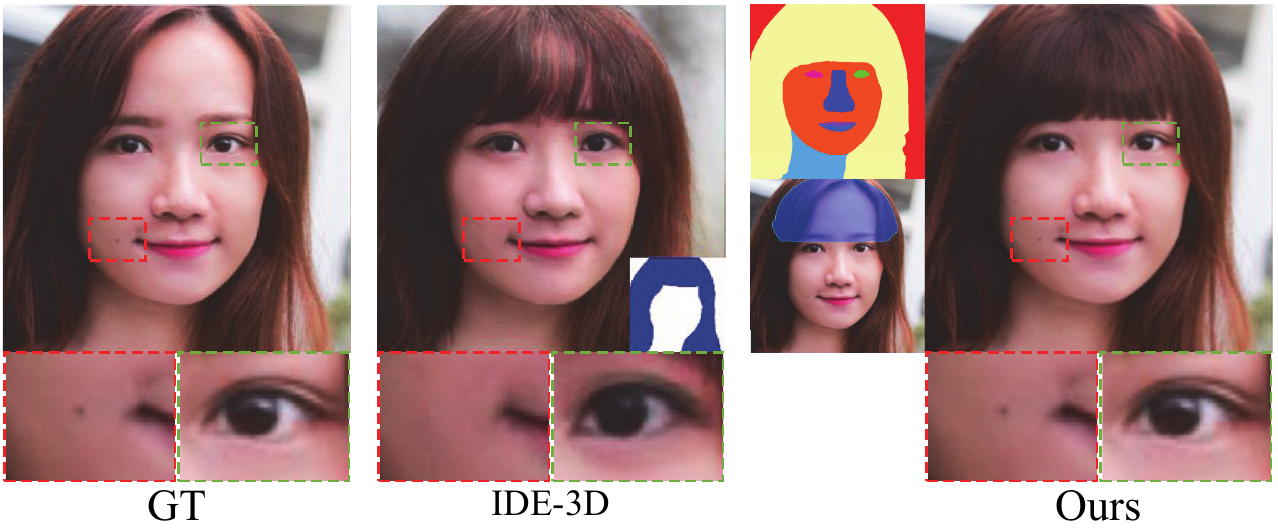}
	\caption{\revise{Visual comparison of the semantic-guided local facial editing with zoom-in details between IDE-3D~\cite{IDE_3D_2022} and FACEMUG. Sub-images are the guidance information. The result of IDE-3D was obtained from the paper~\cite{IDE_3D_2022}.   }
 }\label{fig:fig_ide_3d_compare}
\end{figure}

\begin{figure*}[!htbp]
	\centering
	\includegraphics[width=\textwidth]{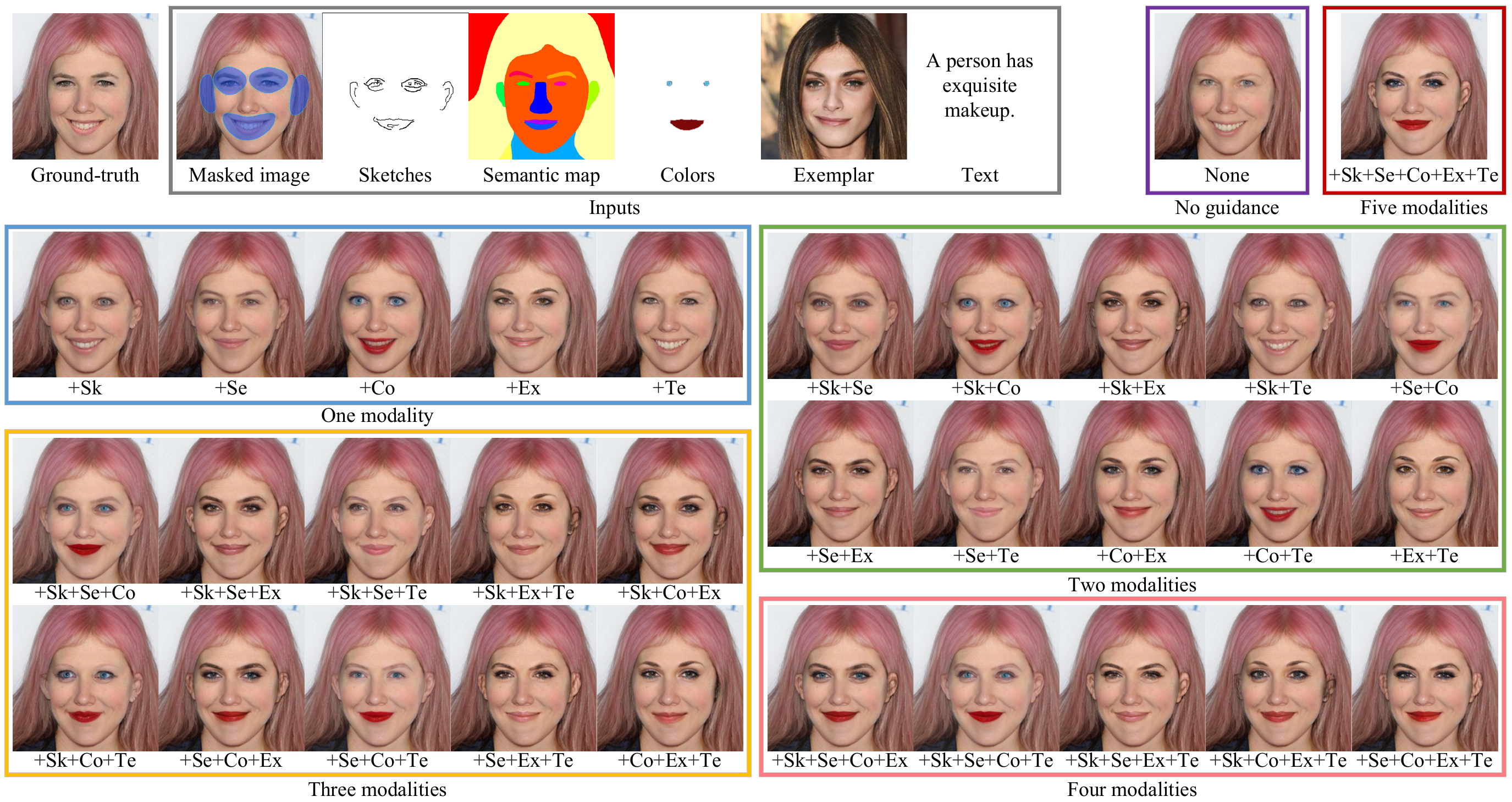}
	\caption{Visual performance of our FACEMUG with various modal inputs. 
    The masked image was utilized for all outputs. There are a total of 32 combinations (subsets) of five modalities.  Our FACEMUG generates visually appealing results and shows high global consistency to the unedited regions.  }\label{fig:fig_effects_multimodal}

	\includegraphics[width=\textwidth]{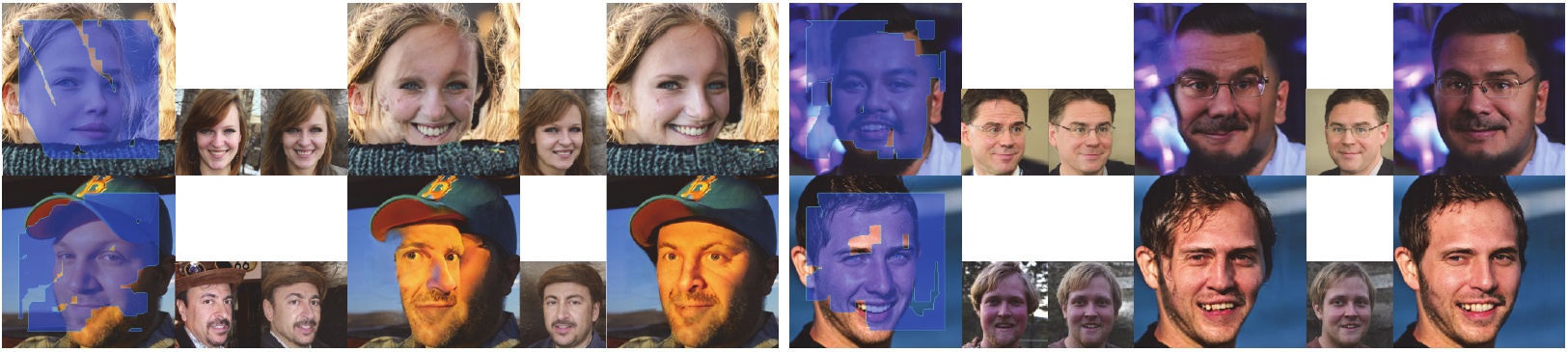}
	\caption{Visual comparison of the ablation study on the latent warping module (from left to right in each group): the masked image, the exemplar image, inversion of latent codes of the exemplar image, the editing result without the latent warping module (+Ex w/o warping), inversion of warped latent codes of the exemplar image, and the editing result with the latent warping module (+Ex). Our warping module improves the visual quality when the pose misalignment happens between the edited image and the exemplar.  }\label{fig:fig_ablation_exemplar}
\end{figure*}


Table~\ref{tab:semantic_generation} presents the quantitative performance of the compared methods. SDM and ColDiffusion achieve good FID scores when conditioned by semantic maps. 
Without colors and unmasked pixels, the performance of FACEMUG (+Se) is limited. Incorporating colors significantly improves the performance of FACEMUG (+Se+Co). With the ability to edit masked regions while preserving unmasked pixels, FACEMUG (+Se+Ma) achieves the best FID, U-IDS, and LPIPS scores for both datasets. FACEMUG outperforms the compared models in terms of the authenticity of locally edited facial images guided by semantic maps.

\revise{
Fig.~\ref{fig:fig_ide_3d_compare} shows the visual comparison between FACEMUG and the SOTA 3D-aware facial editing method, IDE-3D~\cite{IDE_3D_2022}.
Both methods achieve high-quality editing.
IDE-3D supports 3D face synthesis but may result in some losses in facial details during GAN inversion, such as moles and eyelids.
In comparison, FACEMUG excels at preserving unedited facial attributes but faces challenges in reconstructing 3D facial geometry. 
Combining the strengths of both methods could contribute to a more advanced multimodal facial editing framework.
}



\begin{figure*}[!htbp]
	\centering
	\includegraphics[width=\textwidth]{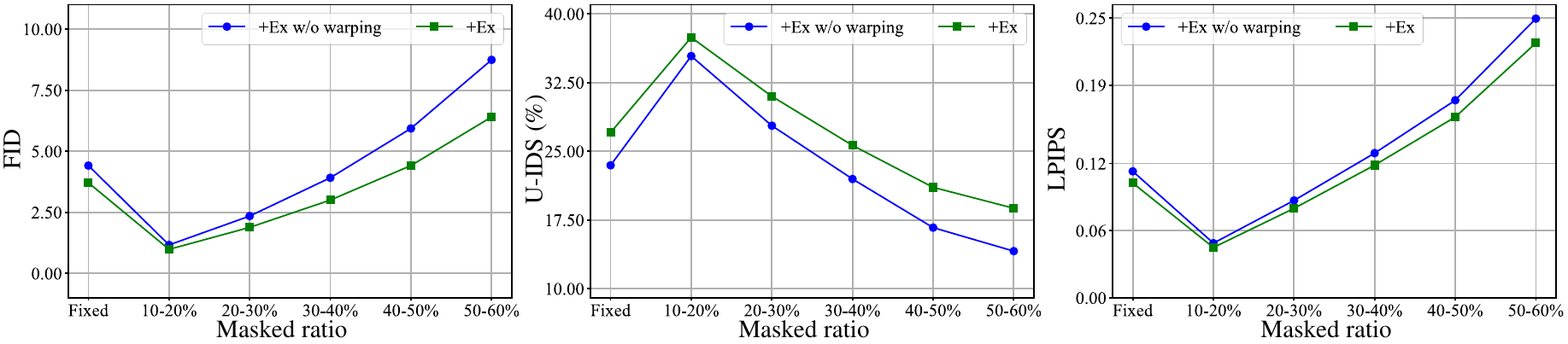}
	\caption{Quantitative comparison of the ablation study on the latent warping module on the FFHQ dataset. ``+Ex w/o warping'': the exemplar image without the latent warping module used; ``+Ex'': the exemplar image with the latent warping module used. Other modalities were not used.	}\label{fig:fig_ffhq_multimodal_ablation_metric}
\end{figure*}

\subsection{Ablation study}

\subsubsection{Ablation study on multimodal inputs}
We also conducted experiments to evaluate the effects of multimodal inputs on our FACEMUG framework. 
Fig.~\ref{fig:fig_effects_multimodal}  showcases editing examples of FACEMUG using a total of 32 input configurations of five modalities. 
Generally, when multimodalities are consistent with each other, incorporating more modalities achieves much better visual quality.
``None'' means that only the masked image is utilized for image inpainting. 
When using sketches or semantic maps, FACEMUG clearly exhibits the structure of inputs. Incorporating colors enhances texture details and produces faithful edited results.
Example images further help transfer the styles and facial identity to the edited regions.
With the guidance of the text, more detailed facial attributes can be manipulated.
Utilizing all modalities (five modalities) information allows FACEMUG to achieve the best overall performance.
It also demonstrates that our method can work well on all the subsets of five modalities.



\subsubsection{Ablation study on latent warping module}
We conducted a study to assess the effectiveness of our latent warping module. 
Various irregular masks with different mask ratios~\cite{Liu2018} and a fixed center 25\% ($128 \times 128$) rectangular mask were used to simulate different editing situations. 
We show visual examples in Fig.~\ref{fig:fig_ablation_exemplar}. We can find that our latent warping module effectively aligns the exemplar pose to the edited image while preserving the facial attributes and identity of the exemplar images.
As shown in Fig.~\ref{fig:fig_ffhq_multimodal_ablation_metric}, the quantitative performance results demonstrate that our latent warping module achieves better quantitative performance (+Ex) compared to not using it (+Ex w/o warping). 
Since the edited and exemplar images contain different poses, directly transferring features to target regions may cause obvious artifacts. Our latent warping module mitigates this issue by adapting the exemplar pose to align with the edited image in the style latent space, thereby avoiding the boundary issue during editing.

\begin{table}[t]
	\caption{Ablation study of the image generation performance of other main components on the FFHQ dataset with $50-60\%$ mask ratios. (A) represents our full model (FACEMUG).  (B) replaces our style fusion blocks with element-wise addition operations. (C) replaces our style fusion blocks with gated convolution blocks~\cite{Yu2019}.  (D) replaces our diversity-enhanced attribute loss with the attribute loss~\cite{lu2022inpainting}. (E) removes the facial feature bank. \textbf{Bold}: top-1 quantity.}
	\label{tab:ablation_FACEMUG}
	\centering
		\begin{tabular}{cccc}
			\toprule
			\makebox[0.2\columnwidth][c]{Method} & \makebox[0.2\columnwidth][c]{FID$^{\downarrow}$} & \makebox[0.2\columnwidth][c]{U-IDS$^{\uparrow}$}& \makebox[0.2\columnwidth][c]{LPIPS$^{\downarrow}$}  \\
			\midrule
			A    &\textbf{3.274}  & \textbf{28.14\%} & \textbf{0.1917}  \\
			B   &  {3.829}  &  {24.05\%}  &  0.2010  \\
			C     & {3.478}   &{26.70\%}& 0.1954  \\
			D   & {3.599}   & {26.44\%}& 0.1991 \\
			E    & 3.725    &  25.26\%  &  0.1971  \\
			\bottomrule
	\end{tabular}
\end{table}

\subsubsection{Ablation study on other main components}

Here, we explored the image generation performance of other main components by comparing FACEMUG to its five variants on the FFHQ dataset with $50-60\%$ mask ratios, as shown in Table~\ref{tab:ablation_FACEMUG}.
We tested to replace our fusion blocks with element-wise addition operations (B) and gated convolution blocks~\cite{Yu2019} (C). The quantitative performance of models (B) and (C)  exhibited a certain degree of decline, compared to our full model (A).
When replacing our diversity-enhanced attribute loss with the attribute loss~\cite{lu2022inpainting} (D), the quantitative measures also show a decrease. 
We also show FACEMUG's effectiveness by removing the facial feature bank (E). The quantitative scores dropped significantly, underscoring the importance of the facial feature bank for high-quality results.

\section{Conclusion}\label{sec:conclusion}
In this paper, we have explored a novel multimodal generative and fusion framework (FACEMUG) for globally-consistent local facial editing, {which generates realistic facial features in response to multimodal inputs on the edited regions while maintaining visual coherence with unedited parts.}
FACEMUG takes advantage of diverse input modalities (e.g., sketches, semantic maps, color maps, exemplar images, text, and attribute labels) to perform fine-grained and semantic facial editing on geometry, color, expressions, attributes, and identity within edited regions, {and allows for incremental edits.}
Extensive experiments have demonstrated the effectiveness of the proposed method.


\subsubsection*{Limitations and future work} Our approach comes with certain limitations. First, although the inference of FACEMUG is fast, the training of FACEMUG is time-intensive and requires approximately one month to complete on a V100 GPU. We plan to develop a more lightweight model to expedite FACEMUG's training process in the future. 
\revise{Second, because of the limited training data, the pre-trained StyleGAN struggles to generate relatively extreme expressions, poses, and appearances, potentially leading to FACEMUG failures in these domains.  A more expressive and powerful pre-trained StyleGAN will improve our model.}
Third, like most multimodal editing algorithms~\cite{huang2022poegan}, FACEMUG may not generate satisfying results when different modalities contain contradictory guiding information. Designing a more sophisticated technique to reweight the contribution of each modality would be a promising step forward. 
\revise{
Finally, an interesting future direction is to incorporate more modalities, such as text and audio, into our multimodal aggregation module to achieve more diverse editing.  A possible solution is to employ facial landmarks as an intermediate motion representation~\cite{hu2023vectortalker}.
}



\appendix

\revise{
\section{{More implementation details}}

\subsection{{Latent warping module}}

\begin{figure}[b]
	\centering
	\includegraphics[width=0.485\textwidth]{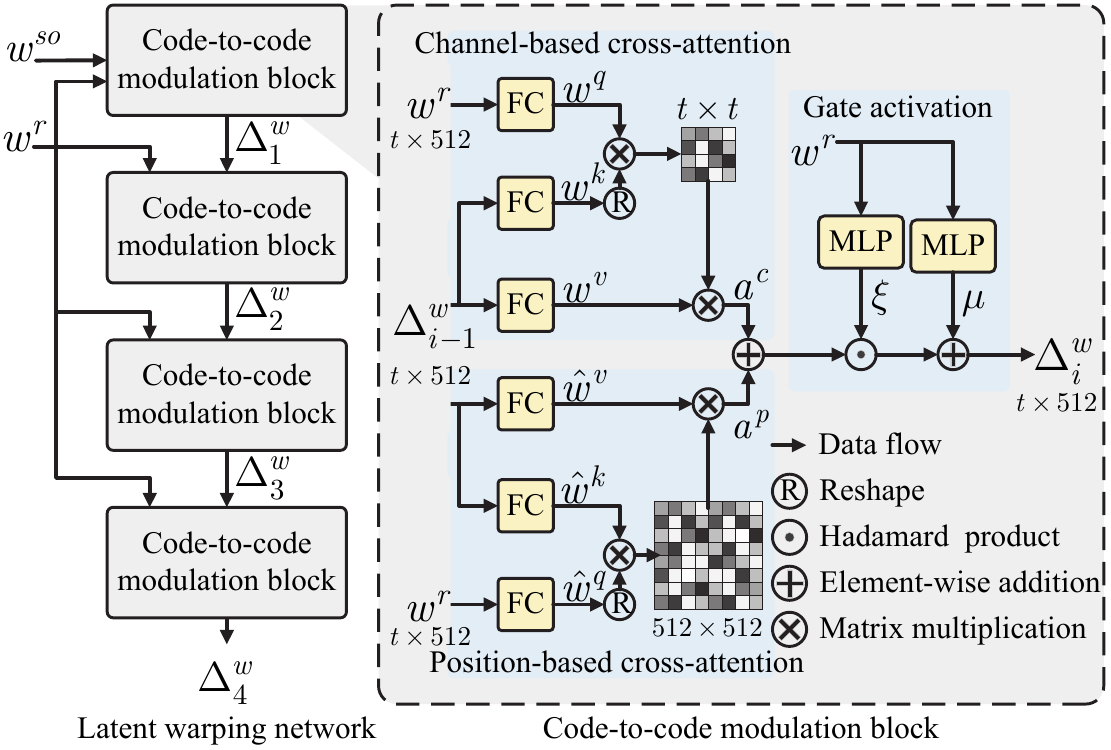}
	\caption{\revise{The architecture of the latent warping network.}}\label{fig:fig_warping_module_net}
\end{figure}

Let $w^{ta} \in \mathbb{R}^{t \times 512 \times 1}$ and $w^{so} \in \mathbb{R}^{t \times 512 \times 1}$ be the target and source latent codes,  and $w^{r}= w^{ta}-w^{so}$ be the residual latent codes between $w^{ta}$ and $w^{so}$.
As illustrated in Fig.~\ref{fig:fig_warping_module_net} (right), each code-to-code modulation block computes channel-based attention~\cite{fu2019dual}, position-based attention~\cite{fu2019dual}, and gated maps~\cite{Yu2019} between $w^{r}$ and $\Delta^w_{i-1}$, and outputs the latent codes $\Delta^w_{i}$ ($i = 1, 2, 3, 4$):
\begin{eqnarray}\label{equ:latent_warping_module}
	\begin{aligned}
		&w^q =\operatorname{FC}(w^{r}), w^k=\operatorname{FC}(\Delta^w_{i-1}), w^v=\operatorname{FC}(\Delta^w_{i-1}), \\
		&a^c =  \operatorname{Softmax}(w^q \cdot w^{k^{\top}}  / \tau_1) \cdot w^v ,\\
		&\hat{w}^q =\operatorname{FC}(w^{r}), \hat{w}^k=\operatorname{FC}(\Delta^w_{i-1}), \hat{w}^v=\operatorname{FC}(\Delta^w_{i-1}), \\
		&a^p = \hat{w}^v \cdot \operatorname{Softmax}( \hat{w}^{q^{\top}} \cdot \hat{w}^k  / \tau_2),\\
		&\xi =\sigma(\operatorname{MLP}(w^{r})), \mu=\phi(\operatorname{MLP}(w^{r})),\\
		&\Delta^w_{i} = \operatorname{LayerNorm}(a^p+a^c) \odot (\xi + 1) + \mu,
	\end{aligned}
\end{eqnarray}
where $w^q$,  $w^k$, $w^v$, $a^c$, $\hat{w}^q$, $\hat{w}^k$, $\hat{w}^v$, $a^p$, $\xi$, $\mu$, and $\Delta^w_{i}$ have the same dimension as ${w}^{so}$, and $\Delta^w_{0} = {w}^{so}$. We set $\tau_1=\sqrt{t}$ and $\tau_2=\sqrt{512}$.
Our code-to-code modulation block calculates query projections ($w^q$ and $\hat{w}^q$), key projections ($w^k$ and $\hat{w}^k$), and value projections ($w^v$ and $\hat{w}^v$) to obtain channel-based and position-based cross-attention maps, respectively. This allows us to obtain the reorganized latent codes $a^c$ and $a^p$. The gated maps $\xi$ and the bias $\mu$ are utilized to assign importance weights and offsets to each element. $\operatorname{FC}(\cdot)$ is a fully connected layer; $\operatorname{Softmax(\cdot)}$ is the softmax activation; $\operatorname{MLP}(\cdot)$ is a stack of two fully connected layers;  $\operatorname{LayerNorm}(\cdot)$ is the LayerNorm layer.

The latent warping network $H_{\theta_{h}}(\cdot)$ outputs the latent codes $\Delta^w_{4}$ and we can obtain the warped latent codes ${w}^{wa} = \Delta^w_{4} + w^{so}$. By leveraging the code-to-code modulation mechanisms, our latent warping module can effectively align the pose of the source latent representations with the target latent codes.

\begin{figure}[b]
	\centering
	\includegraphics[width=0.49\textwidth]{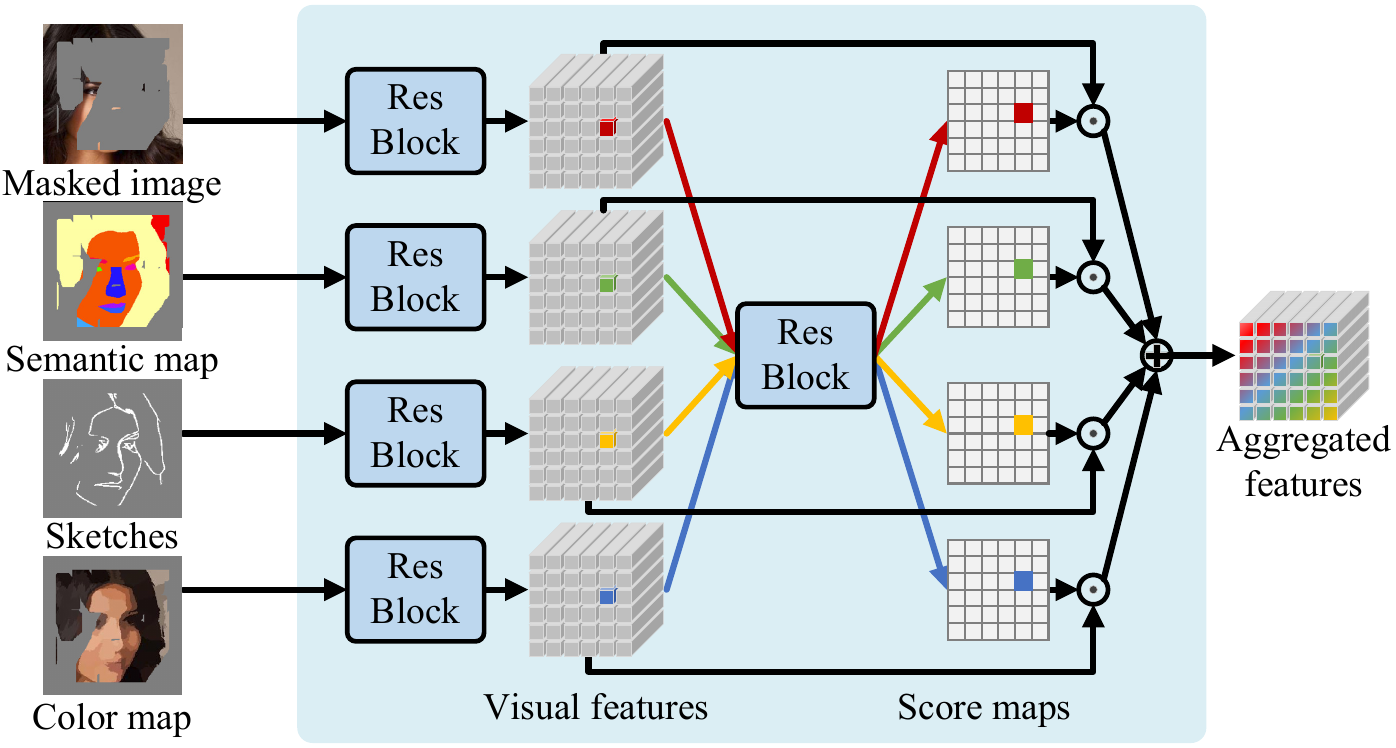}
	\caption{\revise{Illustration of the multimodal aggregation module.}
	}\label{fig:fig_aggregation_module}
\end{figure}

\subsection{{Multimodal aggregation module}}
As shown in Fig.~\ref{fig:fig_aggregation_module}, for the given pixel-wise multimodal inputs $\mathcal{X}=\{\mathbf{I}_{m},\mathbf{I}_1,\mathbf{I}_2, \ldots,\mathbf{I}_n\}$, we first employ a residual block to extract feature maps for each modality,  resulting in a feature set $\{\mathbf{F}_0^{a},\mathbf{F}_1^{a},\mathbf{F}_2^{a}, \ldots,\mathbf{F}_n^{a}\}$, where $\mathbf{F}_j^{a} \in \mathbb{R}^{h \times w  \times c_{a}}$ and $j=0,1,\ldots,n $. 
Subsequently, a shared residual block is utilized to compute the contribution scores for each spatial point across all pixel-wise modalities, producing $n+1$ score maps $\{\mathbf{B}_0,\mathbf{B}_1,\mathbf{B}_2, \ldots,\mathbf{B}_n\}$, where $\mathbf{B}_j \in \mathbb{R}^{h \times w}$ and $j=0,1,\ldots,n $. To obtain the normalized contribution score for each modality, the softmax activation is applied to normalize scores along the channel dimension. Specifically, the normalized score map $\hat{\mathbf{B}}_k \in \mathbb{R}^{h \times w }$ for the $k$-th modality is computed as follows:
\begin{equation}\label{equ:normalized_score)}
		\begin{aligned}
			\hat{\mathbf{B}}(u,v)_k=\frac{\exp \left( \mathbf{B}(u,v)_k \right)}{\sum_{j=0}^n \exp \left( \mathbf{B}(u,v)_j\right)},
		\end{aligned}
\end{equation}
where $(u,v)$ denotes the spatial point. The contribution score map adaptively weights the importance of each modality in a pixel-wise fashion for the aggregation process. This adaptive weighting mechanism allows the model to assign higher importance to the more informative and detailed pixel-wise modalities while reducing the impact of less informative inputs. The final aggregated multimodal feature is obtained with the broadcasting technique:
\begin{equation}\label{equ:aggregation}
		\begin{aligned}
			\hat{\mathbf{F}}^{a} = \sum^{n}_{j=0} \mathbf{F}^{a}_j \odot \hat{\mathbf{B}}_j.
		\end{aligned}
\end{equation}


\subsection{{Training of multimodal aggregation module, refinement auto-encoder and discriminator}}
The identity loss, the LPIPS loss, the diversity-enhanced attribute loss, and the adversarial loss are combined to optimize the multimodal aggregation module, refinement auto-encoder, and discriminator.

\textit{Identity loss.} 
The identity loss  ${\mathcal{L}}_{id}(\mathbf{I}_{out},\mathbf{I}_{ex})$ is employed to constrain the identity similarity between the edited image $\mathbf{I}_{out}$ and the exemplar image $\mathbf{I}_{ex}$.

The identity  loss~\cite{lu2022inpainting,zhu2022one} is defined as:
\begin{equation}\label{equ:id_lpips_losses2}
	\begin{aligned}
		{\mathcal{L}}_{id}(\mathbf{I}_{out},\mathbf{I}_{ex})= 1- \cos(R(\mathbf{I}_{out}),R(\mathbf{I}_{ex})),
	\end{aligned}
\end{equation}
where $R(\cdot)$ is a pre-trained ArcFace network~\cite{Deng2019}.

\textit{LPIPS loss.}\label{sec:LPIPS_loss}
The LPIPS loss  ${\mathcal{L}}_{lpips}(\mathbf{I}_{out},\mathbf{I}_{gt})$ is applied to enforce the perceptual similarity between $\mathbf{I}_{out}$ and $\mathbf{I}_{gt}$.  When $\mathbf{I}_{gt}$ and $\mathbf{I}_{ex}$ are not from the same image, we set ${\mathcal{L}}_{lpips}(\mathbf{I}_{out},\mathbf{I}_{gt})=0$.

The LPIPS loss~\cite{Zhang2018} is defined as:
\begin{equation}\label{equ:warp_lpips_losses1}
	\begin{aligned}
		{\mathcal{L}}_{lpips}(\mathbf{I}_{out},\mathbf{I}_{gt})=\| P(\mathbf{I}_{out})-P(\mathbf{I}_{gt})\|_{2},
	\end{aligned}
\end{equation}
where $P(\cdot)$ corresponds the pre-trained perceptual feature extractor VGG~\cite{Simonyan2014}. 


\textit{Diversity-enhanced attribute loss.}\label{sec:latent_loss}
In order to learn the mapping between style latent codes and corresponding facial attributes, and support attribute-conditional editing in the style latent space, we propose a diversity-enhanced attribute loss to constrain the consistency between facial attributes of the edited image $\mathbf{I}_{out}$ and the interpolated latent codes $\overline{w}$. To emulate the latent editing process during the training phase, our approach involves style mixing and interpolation operations. We first apply the style mixing~\cite{Karras2020} to get mixed latent codes $\hat{w}^z$ from two random latent codes. Then we generate exemplar latent codes ${w}^e= E_{\hat{\theta}_{e}}(\mathbf{I}_{ex})$ and interpolate between $\hat{w}^z$  and  ${w}^e$ to obtain the interpolated latent codes $\overline{w}$:
\begin{equation}\label{equ:mixing_style} 
	\begin{aligned}
		\hat{w}^z=& \operatorname{Mixing}(F_{\hat{\theta}_{f}}(z_1),F_{\hat{\theta}_{f}}(z_2)),\\
		\overline{w}=&  \alpha  \cdot  w^e + (1- \alpha) \cdot \hat{w}^z, \\
	\end{aligned}
\end{equation}
where $z_1 \in \mathcal{Z}  $ and $z_2 \in \mathcal{Z} $ are two random latent codes, $\alpha \in [0,1]$ is the uniformly sampled random number, and we set $\alpha =1.0$ when $\mathbf{I}_{gt}=\mathbf{I}_{ex}$. Then, we modulate the interpolated latent codes  $\overline{w}$ into the multimodal generator to obtain the edited image $\mathbf{I}_{out}$. We then apply the attribute loss to constrain the training in the style latent space, i.e., $\mathcal{L}_{attr}({w}^o,\overline{w})$,
where ${w}^o= E_{\hat{\theta}_{e}}(\mathbf{I}_{out})$.

The attribute loss~\cite{lu2022inpainting,hou2022feat} is defined as:
\begin{equation}\label{equ:loss_warping2}
	\begin{aligned}
		\mathcal{L}_{attr}({w}^{o}, \overline{w})= \| {w}^{o}  - \overline{w} \|_{2}.
	\end{aligned}
\end{equation}

The diversity-enhanced attribute loss is designed to mimic the latent attribute editing process, promoting a diverse range of embedded latent codes throughout training. By exposing the multimodal generator to a wide variety of mapping cases, the model becomes more effective in handling potential latent codes, thereby improving the latent attribute editing capability.

\textit{Adversarial loss.} 
We use the adversarial non-saturating logistic loss~\cite{Goodfellow2014} with $R_{1}$ regularization~\cite{Mescheder2018}:
\begin{equation}\label{equ:loss_adv}
	\begin{aligned}
		\mathcal{L}_{adv}&(\mathbf{I}_{out},\mathbf{I}_{gt}) =\mathbb{E}_{\mathbf{I}_{out}}[\log(1-D(\mathbf{I}_{out})] \\
		&+\mathbb{E}_{\mathbf{I}_{gt}}[\log(D(\mathbf{I}_{gt}))] - \frac{\gamma}{2} \mathbb{E}_{\mathbf{I}_{gt}}[\|  \nabla_{\mathbf{I}_{gt}} D(\mathbf{I}_{gt})\|_{2}^{2}],\\
	\end{aligned}
\end{equation}
where $\gamma  = 10 $ is used to balance the $R_1$ regularization term.


\textit{Total loss.}\label{sec:total_losses2}
The total training loss is defined as:
\begin{equation}\label{equ:loss_total2}
	\begin{aligned}
			{O}(\theta_{a},\theta_{g},&\theta_{d}) =  \lambda_{id}\mathcal{L}_{id}(\mathbf{I}_{out},\mathbf{I}_{ex})+\lambda_{attr}\mathcal{L}_{attr}({w}^o,\overline{w})\\
			&+ \lambda_{lpips}\mathcal{L}_{lpips}(\mathbf{I}_{out},\mathbf{I}_{gt})+\mathcal{L}_{adv}(\mathbf{I}_{out},\mathbf{I}_{gt}),
	\end{aligned}
\end{equation}
where $\lambda_{id}=0.1$, $\lambda_{lpips}=0.5$, and $\lambda_{attr}=0.1$ in this work.

For each iteration, we obtain the optimized parameters ${\theta}^{*}_{a}$, ${\theta}^{*}_{g}$ and ${\theta}^{*}_{d}$ via the minimax game iteratively:
\begin{equation}\label{equ:loss_minimax}
	\begin{aligned}
		({\theta}^{*}_{a},{\theta}^{*}_{g})&=\arg \min _{\theta_{a},\theta_{g}} {O}(\theta_{a},\theta_{g}, \theta_{d}),\\
		{\theta}^{*}_{d}&=\arg \max _{\theta_{d}} {O}(\theta_{a},\theta_{g}, \theta_{d}).
	\end{aligned}
\end{equation}

The refinement network $G_{\theta_g}$ is trained to generate a realistic edited image $\mathbf{I}_{out}$ while the discrinimator $D_{\theta_g}$ tries to  differentiate between $\mathbf{I}_{gt}$ and $\mathbf{I}_{out}$. In an alternating fashion, $A_{\theta_a}$ and $G_{\theta_g}$ are trained in a phase while $D_{\theta_d}$  is trained in the other. Note that  $\hat{\theta}_{f}$, $\hat{\theta}_{e}$, and $\hat{\theta}_{s}$ are keeping unchanged. 
During training, we attempted to embed all pixel-wise modalities in each iteration, but the model tended to overfit with the combined modalities and struggled with missing ones. As a solution, we randomly removed pixels and dropped out some input modalities for each training iteration, enhancing the robustness of missing modalities during inference. 
}

\subsection{Training peseudo-codes}   

The pseudo-codes of our training procedures are provided in \revise{Algorithm~\ref{algo:encoder_training}}, Algorithm~\ref{algo:latent_warping} and Algorithm~\ref{algo:FACEMUG_framework}.
The threshold $\rho \in [0,1]$ is used to control the probability that the sampled ground-truth image and exemplar image are the same. The threshold $\omega \in [0,1]$ and the random masks $\hat{\mathcal{M}}$ are used to control the sparsity for multimodal inputs. We set $\rho=0.5$ and $\omega=0.8$ in this paper. 

\revise{
\begin{algorithm}[!htbp]
	\caption{Training procedure of the style encoder}
	\label{algo:encoder_training}
	\small 
	\begin{algorithmic}[1]
\revise{
		\While {$E_{\theta_{e}}$ have not converged}
		\State Sample batch images $\mathcal{I}_{gt}$ from training data
  		\State Set exemplars from ground-truth $\mathcal{I}_{0} \leftarrow \mathcal{I}_{gt}$
		\State Sample corresponding multimodal inputs $\mathcal{I}_{1},\mathcal{I}_{2},\ldots,\mathcal{I}_{n}$
		\State Create random masks $\mathcal{M}$ 
		\State Sample a random number $r  \in [0,n]$
            \For{$i = 0$ to $n$}
    		\If {$r = i $}
    		      \State Get masked modality $\mathcal{I}_i \leftarrow \mathcal{M} \odot  \mathcal{I}_i $
		      \Else
                    \State Set modality $\mathcal{I}_i$ to be zero tensor
                \EndIf
            \EndFor        
            \State Set inputs $\mathcal{X} \leftarrow \{ \mathcal{I}_{0},\mathcal{I}_{1},\mathcal{I}_{2},\ldots,\mathcal{I}_{n} \}$
            \State Get projected latent codes $w^p \leftarrow  E_{{\theta}_{e}}(\mathcal{X})$
            \State Get a reconstructed image $\mathcal{I}_p \leftarrow  S_{\hat{\theta}_{s}}({w^p})$
	
		\State Update $\theta_{e}$ using the total loss defined in e4e
		\EndWhile
  }
	\end{algorithmic}
\end{algorithm}
}

\begin{algorithm}[!htbp]
	\caption{Training procedure of our latent warping module}
	\label{algo:latent_warping}
	\small
	\begin{algorithmic}[1]
		\While {$H_{\theta_{h}}$ has not converged}
		\State Sample batch images $\mathcal{I}_{gt}$ from training data
		\State Calculate augmented images $\mathcal{I}_{ta}$ from $\mathcal{I}_{gt}$
		\State Calculate mirror flipped images $\mathcal{I}_{f}$ from $\mathcal{I}_{gt}$
		\State Get ground-truth latent codes ${w}^{gt} \leftarrow E_{\hat{\theta}_{e}}(\mathcal{I}_{gt})$ 
		\State Get target latent codes ${w}^{ta} \leftarrow E_{\hat{\theta}_{e}}(\mathcal{I}_{ta})$ 
		\State Get flipped latent codes ${w}^{f} \leftarrow E_{\hat{\theta}_{e}}(\mathcal{I}_f)$
		\State Sample a random number $\beta \in [0,1]$
		\State Get source latent codes ${w}^{so} \leftarrow   \beta  \cdot  w^{gt} + (1- \beta) \cdot {w}^f$
		\State Get results ${w}^{wa} \leftarrow H_{\theta_{h}}(w^{ta}-w^{so},w^{so}) + w^{so}$
		\State Update $\theta_h$ with $\mathcal{L}_{attr}$, ${\mathcal{L}}_{id}$, and ${\mathcal{L}}_{lpips}$ 
		\EndWhile
	\end{algorithmic}
\end{algorithm}

\begin{figure*}[!t]
	\centering
	\includegraphics[width=\textwidth]{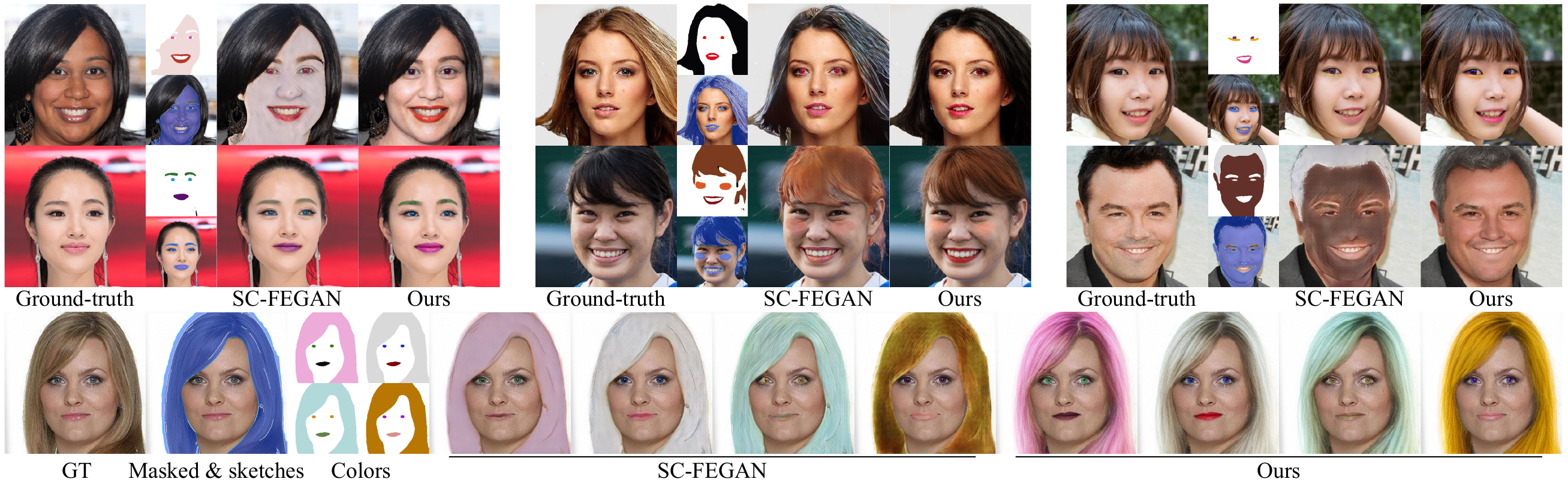}
	\caption{Visual comparison of color-guided facial editing between SC-FEGAN~\cite{Jo2019} and ours: (first two rows) sub-images represent the colors and masked images with sketches; (bottom row) diverse editing results guided by different color maps. Our method produces images with higher quality. }
	\label{fig:fig_makeup_cases}
\end{figure*}

\begin{algorithm}[!htbp]
	\caption{Training procedure of the multimodal aggregation module, refinement auto-encoder, and discriminator}
	\label{algo:FACEMUG_framework}
	\small 
	\begin{algorithmic}[1]
		\While {$A_{\theta_{a}}$, $G_{\theta_{g}}$, and $D_{\theta_{d}}$ have not converged}
		\State Sample batch images $\mathcal{I}_{gt}$ from training data
		\State Sample corresponding multimodal inputs $\mathcal{I}_{1},\mathcal{I}_{2},\ldots,\mathcal{I}_{n}$
		\State Create random masks $\mathcal{M}$ 
		\State Sample a random number $r \in [0,1]$
		\If {$r > $ threshold $\rho$}
		\State Sample exemplars $\mathcal{I}_{ex}$ from training data
		\State Set  $\mathcal{I}_{1},\mathcal{I}_{2},\ldots,\mathcal{I}_{n}$ to be zero tensors
		\Else
		\State Set exemplars from ground-truth $\mathcal{I}_{ex} \leftarrow \mathcal{I}_{gt}$
		\State Sample random numbers $q_1, q_2, \ldots, q_n \in [0,1]$
		\For{$i = 1$ to $n$}
		\If {$q_i > $ threshold $\omega$}
		\State Set modality $\mathcal{I}_i$ to be zero tensor
		\Else
		\State Create random masks $\hat{\mathcal{M}}$
		\State Get masked modality $\mathcal{I}_i \leftarrow \hat{\mathcal{M}} \odot \mathcal{M} \odot  \mathcal{I}_i $
		\EndIf
		\EndFor
		\EndIf
		\State Get masked images $\mathcal{I}_{m} \leftarrow \mathcal{I}_{gt} \odot (1 - \mathcal{M}) $
		\State Set inputs $\mathcal{X} \leftarrow \{ \mathcal{I}_{m},\mathcal{I}_{1},\mathcal{I}_{2},\ldots,\mathcal{I}_{n} \}$
		\State Get exemplar latent codes $w^e \leftarrow  E_{\hat{\theta}_{e}}(\mathcal{I}_{ex})$
		\If {$r > $ threshold $\rho$}
		\State Sample random latent vectors $\mathcal{Z}_1$ and $\mathcal{Z}_2$  
		\State Get mixed codes $\hat{w}^z \leftarrow \operatorname{Mixing}(F_{\theta_{f}}(\mathcal{Z}_1),F_{\theta_{f}}(\mathcal{Z}_2))$ 
		\State Sample a random number $\alpha \in [0,1]$
		\State Get interpolated codes $\overline{w} \leftarrow \alpha  \cdot  w^z + (1- \alpha) \cdot \hat{w}^e$
		\Else
		\State Get codes from exemplar latent codes $\overline{w} \leftarrow w^e$
		\EndIf
		\State Get projected latent codes $w^p \leftarrow  E_{\hat{\theta}_{e}}(\mathcal{X})$
		\State Extract facial features from the StyleGAN generator $\mathcal{F}^s \leftarrow  S_{\hat{\theta}_{s}}({w^p})$
		\State Extract aggregated features $\mathbf{F}^a \leftarrow  A_{{\theta}_{a}}(\mathcal{X})$
		\State Get $\mathcal{I}_{out} \leftarrow \mathcal{I}_{m} \odot (\mathbf{1} - \mathcal{M}) + G_{\theta_g}(\mathbf{F}^a,\overline{w},\mathcal{F}^s) \odot \mathcal{M}$
		\State Update $\theta_{g}$ with  $\mathcal{L}_{adv}$, $\mathcal{L}_{id}$, $\mathcal{L}_{lpips}$, and $\mathcal{L}_{attr}$
		\State Update $\theta_{d}$ with $\mathcal{L}_{adv}$
		\EndWhile
	\end{algorithmic}
\end{algorithm}

\section{More experimental results and comparisons}

\subsection{Comparison on color-guided facial editing}   
We compared FACEMUG to SC-FEGAN~\cite{Jo2019} on color-guided facial editing using sketches and colors. We masked out the edited pixels and utilized the corresponding color information to guide makeup generation. To preserve the source facial structures and other facial attributes, we extracted sketches using a Canny edge detector to extract sketches for guiding the generation of facial geometry. Fig.~\ref{fig:fig_makeup_cases} presents visual illustrations of editing results. Results from SC-FEGAN show clear responses to the input sketches and colors. However, some visual artifacts are noticeable in the masked regions. In contrast, FACEMUG utilizes sketches for facial geometry and colors for color editing to enable customized makeup editing within editing regions.  Our style fusion blocks efficiently utilize extracted features to produce more visually appealing results.

\subsection{Comparison on local attribute-conditional facial editing}

Through aligning our latent space with the $\mathcal{W}+$ style latent space~\cite{Karras2020}, we demonstrate that our approach is not only capable of performing attribute-conditional facial edits using off-the-shelf latent-based semantic editing techniques~\cite{shen2020interfacegan,Abdal_StyleFlow2021,harkonen2020ganspace,CLIP2StyleGAN2022}, but also retains the integrity of unedited regions.

\begin{figure*}[t]
	\centering
	\includegraphics[width=\textwidth]{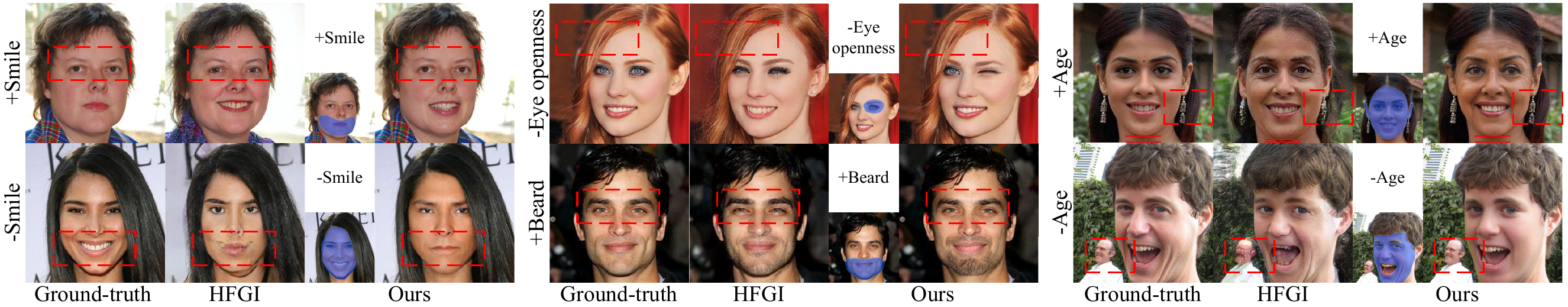}
	\caption{Visual comparison of attribute-conditional editing between HFGI~\cite{wang2022HFGI} and our FACEMUG.  For FACEMUG, the left sub-images in each group represent the attribute label (top) and editing regions (bottom). FACEMUG shows more flexible attribute-conditional editing by only manipulating selected regions (e.g., first row, middle, wink expression), and keeps unrelated features unchanged to show better visual quality and global consistency.  }
	\label{fig:fig_latent_editing_qualitative}
\end{figure*}

Fig.~\ref{fig:fig_latent_editing_qualitative} shows how our method can carry out attribute-conditioned semantic modifications on masked relevant semantic areas. The compared method HFGI~\cite{wang2022HFGI} alters unwanted facial attributes or background information, while our approach ensures the background information (unmasked area) remains unchanged. Additionally, FACEMUG exhibits the capability to execute more complex edits, such as producing a winking expression, attributable to its flexibility in choosing editing regions. 


\subsection{Comparison on exemplar-guided facial editing}\label{sec:exe_guidance}
We also show the comparison with StyleMapGAN~\cite{kim2021stylemapgan},  ILVR~\cite{Choi2021ILVRCM}, and SemanticStyleGAN~\cite{shi2021SemanticStyleGAN} for exemplar-guided editing. To ensure proper facial alignment between the exemplar image and the input image, we extracted the roll, pitch, and yaw angles from the CelebAMask-HQ~\cite{Lee2020} dataset. Subsequently, we selected $550$ pairs with similar poses from the testing set of CelebA-HQ. For each pair, we alternately used one facial image as the exemplar and the another as the masked input image to perform editing. The masked regions were replaced with corresponding exemplar facial features. Publicly available trained models were utilized to generate the editing results.

\begin{figure}[t]
	\centering
	\includegraphics[width=0.45\textwidth]{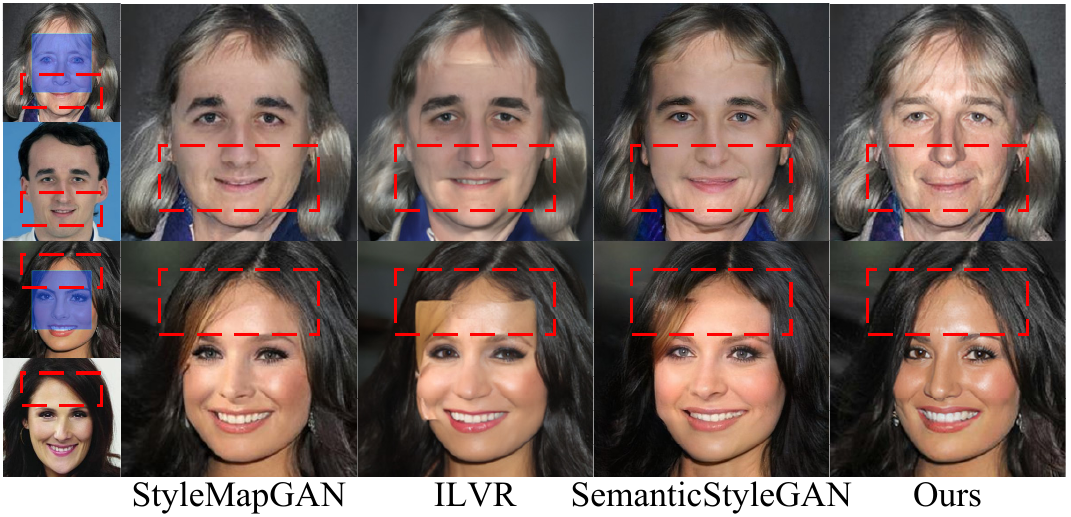}
	\caption{Visual comparison of our FACEMUG to the SOTA exemplar-guided editing methods (StyleMapGAN~\cite{kim2021stylemapgan}, ILVR~\cite{Choi2021ILVRCM}, and SemanticStyleGAN~\cite{shi2021SemanticStyleGAN}): top-left (masked image), bottom-left (exemplar image), right (results).  FACEMUG shows seamless incorporation of exemplar facial attributes while keeping unmasked regions unchanged and achieving global consistency. }
	\label{fig:fig_exemplar_guided_qualitative}
\end{figure}

As shown in Fig.~\ref{fig:fig_exemplar_guided_qualitative}, although most methods produce plausible results, some visible boundary inconsistencies can be found in the details of SemanticStyleGAN, ILVR, and StyleMapGAN. Moreover, these compared methods may introduce unwanted changes in the background or unedited regions. Our FACEMUG can seamlessly fill in the masked pixels using exemplar-like attributes without changing unmasked areas,  yielding high-quality editing results while avoiding the above artifacts.

Table~\ref{tab:exemplar_editing} shows the quantitative performance of compared methods. It shows that exemplar-guided editing is still a challenging task. For SemanticStyleGAN and ILVR, directly filling the corresponding exemplar's facial features into the masked regions may cause obvious artifacts. The style maps in StyleMapGAN can help achieve more harmonious editing results, but it still shows limits on exemplar-guided editing. Our FACEMUG achieves the best FID and LPIPS scores, indicating that the edited images from FACEMUG obtain the highest visual quality.

\begin{table}[t]
	\caption{Quantitative comparison of our method to existing exemplar-guided editing. \textbf{Bold}: top-1 quantity.}
	\label{tab:exemplar_editing}
	\centering
	 \resizebox{\linewidth}{!}{
		\begin{tabular}{ccccc}
			\toprule
			\multicolumn{1}{c}{Metric} & \multicolumn{1}{c}{StyleMapGAN~\cite{kim2021stylemapgan}} & \multicolumn{1}{c}{ILVR~\cite{Choi2021ILVRCM}} & \multicolumn{1}{c}{SemanticStyleGAN~\cite{shi2021SemanticStyleGAN}}& \multicolumn{1}{c}{Ours} \\
			\midrule
			FID$^{\downarrow}$ &   24.59  &   62.27   &32.54 &   \textbf{13.28}\\
			LPIPS$^{\downarrow}$&   0.5833   & 0.3495 &0.5801  & \textbf{0.0851}  \\
			\bottomrule
		\end{tabular}
	 }
\end{table}

\begin{figure*}[!t]
	\centering
	\includegraphics[width=\textwidth]{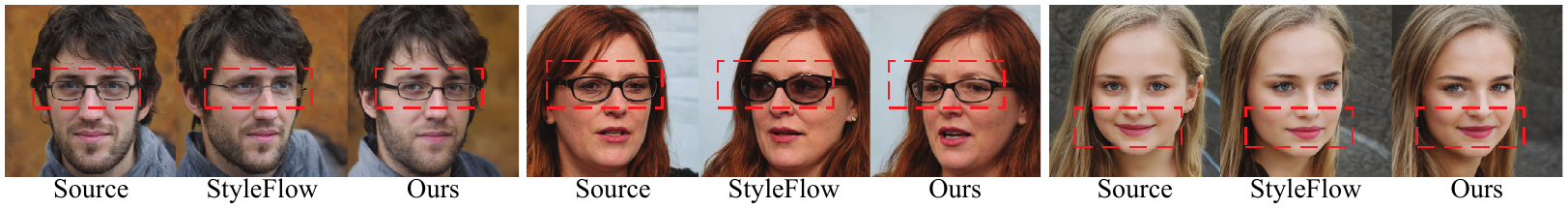}
	\caption{Visual comparison to StyleFlow~\cite{Abdal_StyleFlow2021} on pose editing. Our warping module shows better visual quality and visual consistency for facial attributes.
	}\label{fig:fig_pose_styleflow_qualitative}  
\end{figure*}

\begin{figure*}[!t]
	\centering
	\includegraphics[width=\textwidth]{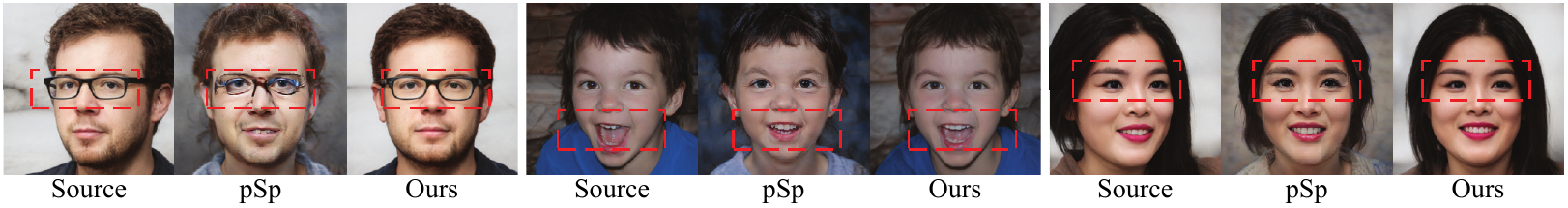}
	\caption{Visual comparison to pSp~\cite{Richardson2021} on face frontalization. Our warping module provides superior visual quality and maintains the consistency of attributes.
	}\label{fig:fig_pose_psp_qualitative}  
\end{figure*}

\begin{figure*}[!t]
	\centering
	\includegraphics[width=\textwidth]{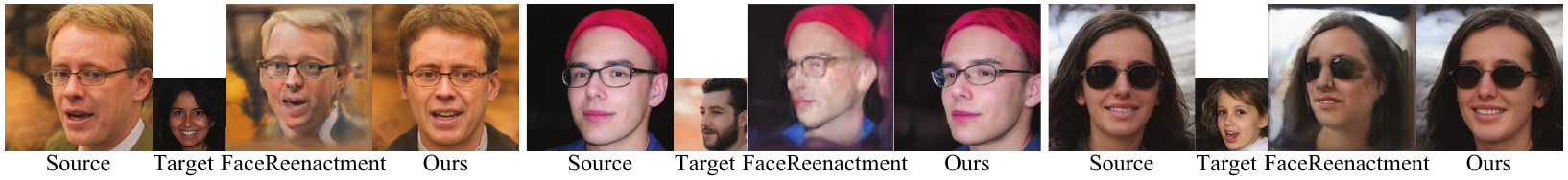}
	\caption{Visual comparison to FaceReenactment~\cite{bounareli2022finding} on pose transfer.	Our warping module ensures enhanced visual quality and consistency of facial features. }\label{fig:fig_pose_reenactment_qualitative}  
\end{figure*}

\begin{figure*}[t]
	\includegraphics[width=\textwidth]{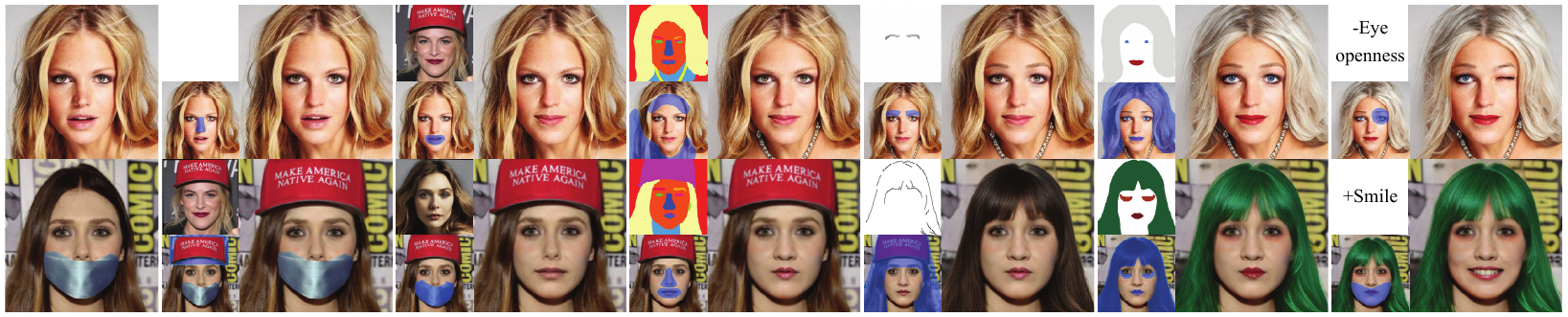}
	\caption{Incremental local facial editing examples with our FACEMUG. Each row: given an input image (first row), FACEMUG incrementally edits the facial image with blemish removal, exemplar-guided facial style transfer, semantic-guided attribute edits, sketch-guided hairstyle edits, color-guided makeup, and attribute-conditioned semantic edits (e.g., gender, age, and expression). For each group, FACEMUG only edits the masked area (bottom-left) guided by the guidance information (top-left) to produce the edited image (right). In the last row's first edit, we copy the hat to the input image and regenerate boundaries seamlessly. }
	\label{fig:teaser_unimodal}
\end{figure*}
\begin{figure*}[t]
	\includegraphics[width=\textwidth]{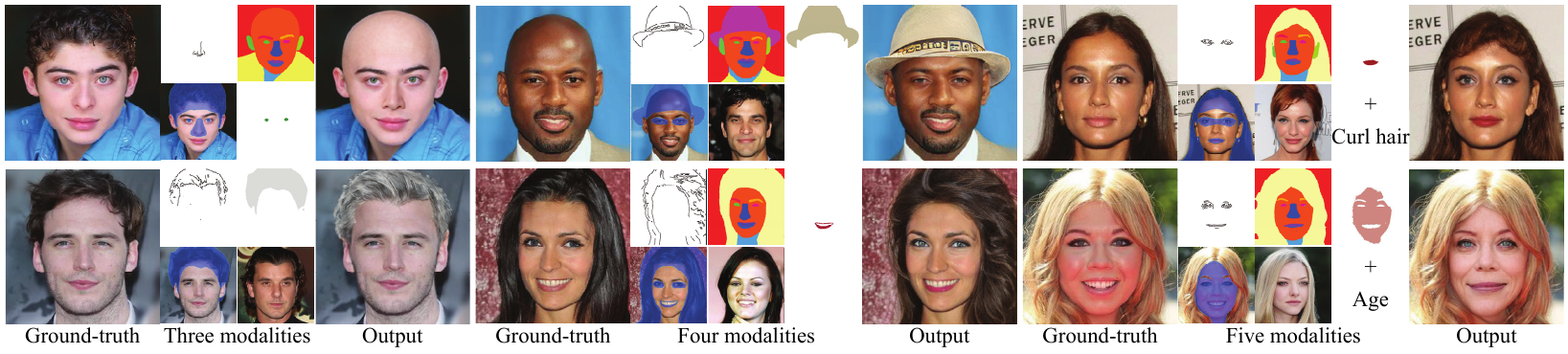}
	\caption{More multimodal local facial editing examples with our FACEMUG. Sketches, semantic maps, colors, exemplars, and attribute labels were utilized for these editing results.
    }
	\label{fig:fig_multi_edits_more}
\end{figure*}


\begin{figure*}[t]
	\centering
	\includegraphics[width=\textwidth]{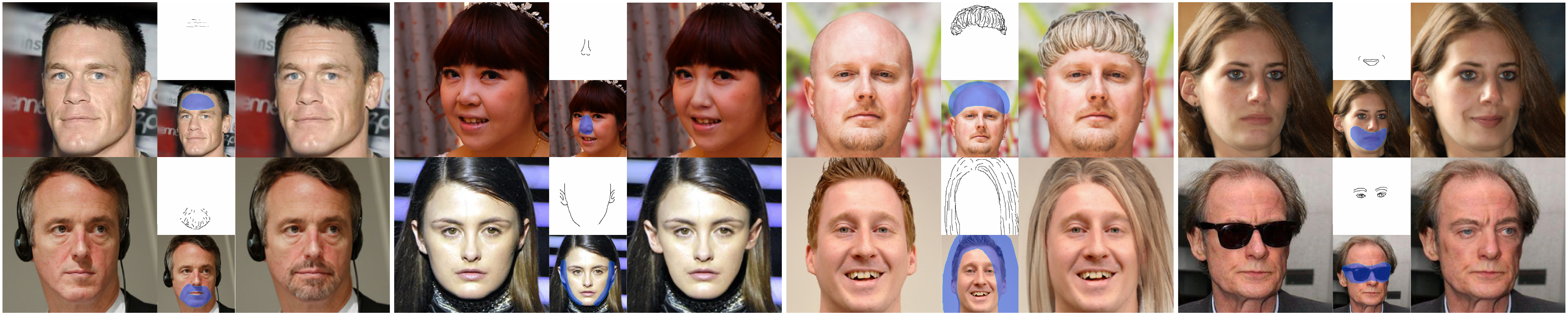}
	\caption{\revise{Examples of sketch-guided facial attribute editing with FACEMUG: texture editing, structure editing, hairstyle editing, and expression editing. For each group:  (left) ground-truth, (top-middle) free-hand sketches, (bottom-middle) masked image, and (right) editing result.}}\label{fig:fig_sketch_editing_cases}  
\end{figure*}

\begin{figure*}[t]
	\centering
	\includegraphics[width=\textwidth]{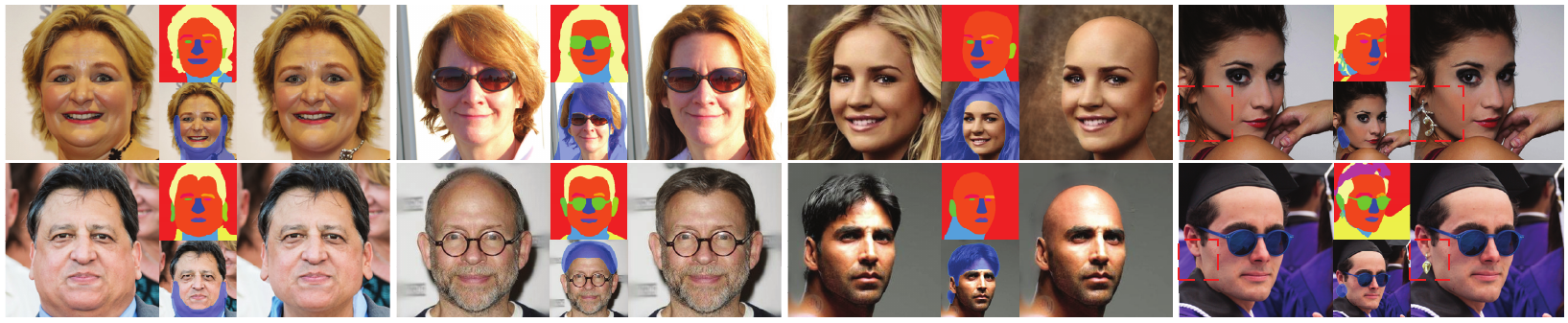}
	\caption{Examples of semantic-guided facial attribute editing with FACEMUG: chin editing, hairstyle editing, hair removal, and accessory addition. For each group:  (left) ground-truth, (top-middle) hand-edited semantic maps, (bottom-middle) masked image, and (right) editing result.   }\label{fig:fig_semantic_editing_cases}  
\end{figure*}

\subsection{Comparison on guided facial pose editing}


To further demonstrate the effectiveness of our latent warping module, we conducted comparisons with SOTA latent-based facial pose editing methods, including StyleFlow~\cite{Abdal_StyleFlow2021}, pSp~\cite{Richardson2021}, and FaceReenactment~\cite{bounareli2022finding} on pose editing, face frontalization, and pose transfer, respectively. We utilized their pre-trained models and codes obtained from their official websites. 
Fig.~\ref{fig:fig_pose_styleflow_qualitative} shows visual examples of editing cases from StyleFlow and ours. StyleFlow excels in editing facial orientation while preserving identity and other facial attributes. However, we observed minor changes in decorative attributes and expressions. In contrast, our latent warping module achieves more natural editing, ensuring consistency in facial attributes through our self-supervised training. 
Fig.~\ref{fig:fig_pose_psp_qualitative} shows that pSp demonstrates face frontalization capabilities but exhibits minor visual artifacts in the generated results. On the contrary,  our latent warping module produces more pleasing outcomes. Moreover, we computed the cosine similarity (CSIM~\cite{bounareli2022finding}) of ArcFace between frontalized images and the ground-truth images for each method on prepared image pairs from Subsection~\ref{sec:exe_guidance}. Compared to pSp with the CSIM score of $0.035$, our FACEMUG achieves the CSIM score of $0.835$, demonstrating the superiority of our method in preserving facial identity.
Fig.~\ref{fig:fig_pose_reenactment_qualitative} shows that FaceReenactment focuses on transferring the facial pose from a target image to a source face. Compared to FaceReenactment, our warping module effectively transfers the facial pose of source images to the target faces with high fidelity.

\subsection{More results on multimodal facial editing}

Here, we show more results guided by multimodalities, including incremental local facial editing in Fig.~\ref{fig:teaser_unimodal} and multimodal local facial editing in Fig.~\ref{fig:fig_multi_edits_more}.
\revise{We also show more facial structure editing of FACEMUG by utilizing free-hand sketches and hand-edited semantic maps. 
As illustrated in Fig.~\ref{fig:fig_sketch_editing_cases}, our FACEMUG framework allows for facial attribute editing using sketches. When provided with masked images and free-hand facial sketches, our method is capable of performing editing on various facial features such as 
texture (wrinkles and beards), structure (nose and chin), hairstyles, and expressions (mouth and eyes) while preserving the unedited regions unchanged.}
As displayed in Fig.~\ref{fig:fig_semantic_editing_cases}, FACEMUG endows users with the capability to edit chins, remove hair, modify hairstyles, and add accessories, all utilizing the provided masked images and semantic maps.
Our approach generates visually appealing and globally consistent images, effectively responding to multimodal inputs while preserving the unmasked areas.

\ifCLASSOPTIONcaptionsoff
  \newpage
\fi



\bibliographystyle{IEEEtran}
\bibliography{IEEEabrv,FACEMUG_bib}

\end{document}